%% file: main.tex
\documentclass{article}

\usepackage[utf8]{inputenc} %
\usepackage[T1]{fontenc}    %
\usepackage{hyperref}       %
\usepackage{url}            %
\usepackage{booktabs}       %
\usepackage{amsfonts}       %
\usepackage{nicefrac}       %
\usepackage{microtype}      %
\usepackage{xcolor}         %

\usepackage{graphicx}
\usepackage{easyReview}
\usepackage{caption}
\usepackage{subcaption}
\usepackage{amsmath}
\usepackage{amsfonts}
\usepackage{amsthm}
\usepackage{amssymb}
\usepackage{bbm}
\usepackage{bm}
\usepackage{lipsum}
\usepackage[numbers]{natbib}
\usepackage{stackengine} %
\usepackage{tabularx}
\usepackage{makecell}
\usepackage{booktabs}
\usepackage{diagbox}
\usepackage{enumitem}

\usepackage{wrapfig,lipsum,booktabs}

\usepackage{pgfplots}
\pgfplotsset{compat = 1.9}
\pgfkeys{/pgf/number format/.cd,
         1000 sep={},
         /pgf/number format/fixed,
         /pgf/number format/precision=5 
}
\usepackage{xcolor}
\definecolor{darkgreen}{RGB}{9, 133, 9}
\usepgfplotslibrary{external}
\usepackage{mathtools}

\usepackage{optidef}
\usepackage{mathtools}

\usepackage{xfrac}

\usepackage{multirow}%
\usepackage{hhline}%

\usetikzlibrary{decorations.pathreplacing,calligraphy} %

\usepackage{bigints}

\usepackage{relsize,exscale}

\usepackage{array}

\usepackage{amsmath}
\usepackage{amssymb}
\usepackage{amsthm}
\usepackage{arydshln}
\usepackage{lipsum}
\usepackage[capitalize,noabbrev]{cleveref}

\usepackage[symbol]{footmisc}

\theoremstyle{plain}

\usepackage[accepted]{icml2024}

\input{0-Preamble}

\icmltitlerunning{Learning in Feature Spaces via Coupled Covariances}

\begin{document}

\twocolumn[

\icmltitle{Learning in Feature Spaces via Coupled Covariances: \\
Asymmetric Kernel SVD and Nystr\"om method}

\icmlsetsymbol{equal}{*}

\begin{icmlauthorlist}
\icmlauthor{Qinghua Tao}{equal,esat}
\icmlauthor{Francesco Tonin}{equal,epfl}
\icmlauthor{Alex Lambert}{esat}
\icmlauthor{Yingyi Chen}{esat}
\icmlauthor{Panagiotis Patrinos}{esat}
\icmlauthor{Johan A.K. Suykens}{esat}
\end{icmlauthorlist}

\icmlaffiliation{esat}{ESAT-STADIUS, KU Leuven, Belgium}
\icmlaffiliation{epfl}{LIONS, EPFL, Switzerland (most of the work was done at ESAT-STADIUS, KU Leuven)}

\icmlcorrespondingauthor{Qinghua Tao, Francesco Tonin}{qinghua.tao@esat.kuleuven.be, francesco.tonin@epfl.ch}

\icmlkeywords{kernel svd, feature maps, asymmetry, covariances}

\vskip 0.3in
]
\printAffiliationsAndNotice{\icmlEqualContribution} 

\begin{abstract}
In contrast with Mercer kernel-based approaches as used e.g., in Kernel Principal Component Analysis (KPCA), it was previously shown that Singular Value Decomposition (SVD) inherently relates to asymmetric kernels and asymmetric Kernel Singular Value Decomposition (KSVD) has been proposed. However, the existing formulation to KSVD cannot work with infinite-dimensional feature mappings, the variational objective can be unbounded, and needs further numerical evaluation and exploration towards machine learning.
In this work, \emph{i)} we introduce a new asymmetric learning paradigm based on coupled covariance eigenproblem (CCE) through covariance operators, allowing infinite-dimensional
feature maps.
{The solution to CCE is ultimately obtained from the SVD of the induced asymmetric kernel matrix, providing links to \ksvd.}
\emph{ii)} Starting from the integral equations corresponding to a pair of coupled adjoint eigenfunctions, we formalize the asymmetric \nystrom method through a finite sample approximation to speed up training.
\emph{iii)} We provide the first empirical evaluations verifying the practical utility and benefits of \ksvd and compare with methods resorting to symmetrization or linear SVD across multiple tasks.\footnote[3]{This work has been accepted at the 41st International Conference on Machine Learning (ICML), 2024. The previous preprint version can be found at \url{https://arxiv.org/abs/2306.07040} and contains useful discussions and insights on KSVD.}
\end{abstract}

\input{1-Introduction}

\input{3-Background}

\input{4-Method}

\input{5-Nystrom}

\input{6-Experiments}

\input{7-Conclusion}

\section*{Acknowledgements}
\label{sec:ack}
This work is jointly supported by ERC Advanced Grant E-DUALITY (787960), iBOF project Tensor Tools for Taming the Curse (3E221427), KU Leuven Grant CoE PFV/10/002, and Grant  FWO G0A4917N, EU H2020 ICT-48 Network TAILOR (Foundations of Trustworthy AI - Integrating Reasoning, Learning and Optimization), and Leuven.AI Institute. This work was also supported by the Research Foundation Flanders (FWO) research projects G086518N, G086318N, and G0A0920N; Fonds de la Recherche Scientifique — FNRS and the Fonds Wetenschappelijk Onderzoek — Vlaanderen under EOS Project No. 30468160 (SeLMA). 
We thank the anonymous reviewers for constructive comments.

\section*{Impact Statement}
This paper presents work whose goal is to advance the field of Machine Learning. There are many potential societal consequences of our work, none of which we feel must be specifically highlighted here.

\bibliography{bibfile}
\bibliographystyle{icml2024}

\newpage
\appendix
\onecolumn
\input{9-Appendix}

\end{document}

%% file: 0-Preamble.tex
\usepackage[utf8]{inputenc} 
\usepackage[T1]{fontenc}    
\usepackage{hyperref}      
\usepackage{url}            
\usepackage{booktabs}       
\usepackage{amsfonts}       
\usepackage{nicefrac}       
\usepackage{xcolor}

\usepackage{amsmath}
\usepackage{amsfonts}
\usepackage{amsthm}
\usepackage{amssymb}
\usepackage{bbm}
\usepackage{bm}
\usepackage{lipsum}
\usepackage{stackengine} 
\usepackage{tabularx}
\usepackage{makecell}
\usepackage{booktabs}
\usepackage{diagbox}
\usepackage{enumitem}

\usepackage{wrapfig,lipsum,booktabs}

\usepackage{pgfplots}
\pgfplotsset{compat = 1.9}
\pgfkeys{/pgf/number format/.cd,
         1000 sep={},
         /pgf/number format/fixed,
         /pgf/number format/precision=5 
}
\usepackage{xcolor}
\definecolor{darkgreen}{RGB}{9, 133, 9}

\usepackage{mathtools}

\usepackage{optidef}
\usepackage{mathtools}

\usepackage{xfrac}

\usepackage{multirow}
\usepackage{hhline}

\usetikzlibrary{decorations.pathreplacing,calligraphy} 

\usepackage{bigints}

\usepackage{relsize,exscale}

\newcommand{\norm}[1]{\left\lVert#1\right\rVert}

\DeclareMathOperator{\fro}{F}

\usepackage{todonotes}
\usepackage{soul}
\newcommand\qh[1]{\textcolor{black}{#1}}
\newcommand\fra[1]{\textcolor{black}{#1}}
\usepackage{xspace}
\newcommand{\ksvd}{KSVD\xspace}
\newcommand{\nystrom}{Nystr\"om\xspace}

\usepackage{array}
\newcolumntype{H}{>{\setbox0=\hbox\bgroup}c<{\egroup}@{}}

\usepackage{amsmath}
\usepackage{amssymb}

\usepackage{amsthm}
\usepackage{arydshln}
\usepackage{lipsum}

\usepackage[capitalize,noabbrev]{cleveref}

\theoremstyle{plain}
\newtheorem{theorem}{Theorem}[section]
\newtheorem{proposition}[theorem]{Proposition}

\theoremstyle{definition}
\newtheorem{definition}[theorem]{Definition}

\theoremstyle{remark}
\newtheorem{remark}[theorem]{Remark}
\Crefname{proposition}{Proposition}{Propositions}
\crefname{problem}{Problem}{Problems}
\Crefname{lemma}{Lemma}{Lemmas}
\Crefname{remark}{Remark}{Remarks}

\newcommand{\brown}[1]{\textcolor{brown}{#1}}
\newcommand{\teal}[1]{\textcolor{teal}{#1}}
\newcommand{\red}[1]{\textcolor[HTML]{f74634}{#1}}

\newcommand{\eg}{{\em e.g.~}}

\newcommand{\cH}{\mathcal{H}}

\newcommand{\cL}{\mathcal{L}}

\newcommand{\cX}{\mathcal{X}}

\newcommand{\cZ}{\mathcal{Z}}

\newcommand\R{\mathbb{R}}

\DeclareMathOperator{\Span}{Span}

\newcommand{\scalone}[1]{\langle \rangle}

\def\fro{\mathrm{F}}

%% file: 1-Introduction.tex
\section{Introduction}
\label{sec:intro}
Feature mappings can transport the data in a Hilbert space of a typically higher dimension.
They are intimately linked through inner products with reproducing kernels \citep{aronszajn1950theory} and thus often associated with symmetric learning.
One can for example think of kernel principal components analysis (KPCA, \citet{scholkopf1998}) where one tries to find orthonormal directions in the feature space that maximize the variance associated to a symmetric Gram matrix, or kernel canonical correlation analysis (KCCA, \citet{lai2000kernelcca}) where the maximization of a correlation based on two different views of the data leads to an optimization problem governed by two symmetric Gram matrices.

In many real-world applications however, there is an inherent degree of asymmetry.
Among others, directed graphs of citation networks \citep{ou2016asymmetric}, biclustering \citep{kluger2003spectral}, attention in Transformers \citep{wright2021,chen2023primal} typically involve an asymmetry that cannot be captured when working with reproducing kernels. 
Often the asymmetric matrices are first symmetrized before applying some matrix decomposition such as singular value decomposition (SVD, \citet{strang2006linear,golub2013matrix}) so that only one set of eigenvectors is obtained.

As a fundamental linear algebra tool, SVD can process arbitrary non-symmetric matrices and jointly learns both left and right singular vectors, e.g., embeddings of source and target nodes \citep{estrada2012structure}.
However, SVD alone lacks flexibility for nonlinear feature learning.
\citet{suykens2016svd} propose asymmetric kernel SVD (\ksvd), a variational principle based on least-square support vector machines (LSSVMs) that leads to the matrix SVD and mentions that nonlinear extensions can be obtained when the SVD is applied to an asymmetric kernel matrix rather than the given data matrix.
However, their formulation only allows finite-dimensional feature mappings to induce the kernel and its variational objective is unbounded unless the regularization hyperparameters are properly selected. 
\qh{Yet, \citet{suykens2016svd} does not provide numerical evaluations on the practical utility and applications of \ksvd,} leaving this topic largely unexplored. \qh{While infinite-dimensional feature maps are common in all kernel methods, including the asymmetric ones, e.g., \citet{wright2021} focus on the understandings of the asymmetric dot-product attention kernel resulting from the queries and keys through a pair of Banach spaces in the supervised setting, little literature addresses learning with generic asymmetric kernel machines with infinite-dimensional maps. Differently, our work provides a new asymmetric learning paradigm for unsupervised feature learning based on the CCE allowing two generic datasets.}

Kernel methods additionally suffer from efficiency, as they require processing a kernel matrix that is quadratic in the sample size.
Many approaches have been proposed to improve the efficiency, among which the Nystr\"om method has been widely applied \citep{NIPS2000_19de10ad, zhang2008improved,gittens2016revisiting,meanti2020kernel,xiong2021nystromformer}.
The  Nystr\"om method of subsampling arises from the approximate eigendecomposition of an integral operator associated with a symmetric kernel \citep{bakerprenter1981numerical},
which restricts the existing \nystrom method to only Mercer kernels.
In \citep{drineas2005nystrom,nystromsvdnemtsov2016matrix,xiong2021nystromformer}, \nystrom-like methods for matrix compression or approximation are discussed by directly applying the symmetric \nystrom method to estimate left and right singular vectors,
yet ignoring the asymmetry constraints. \qh{In \cite{nkcca}, though the asymmetric \nystrom method is mentioned in the proposed nonparametric KCCA method, it still leverages the existing symmetric \nystrom method in implementation for the eigenvectors of two symmetric positive definite kernels and can only deal with square matrices.}
Hence, the analytical framework of the \nystrom method to asymmetric kernel machines remains to be formalized and is of particular interest for the efficient computation of \ksvd.

The research question that we tackle in this paper is ``\emph{How can we learn directions in the feature space in an asymmetric way while controlling the computational complexity of our method?}''

The technical contributions of this work are summarized as:
\begin{itemize}
    \item We first present a new asymmetric learning paradigm based on \emph{coupled covariances eigenproblem} (CCE) allowing infinite-dimensional feature maps.
    We show that its solution leads to the \ksvd problem associated with a specific asymmetric similarity matrix that blends in two feature maps.
    \item We leverage the integral equations involving the pair of adjoint eigenfunctions related to the continuous analog of SVD and derive an extension to the Nystr\"om method able to handle asymmetric kernels, which can be used to speed up \ksvd training without significant decrease in accuracy of the solution.
    \item We conduct extensive experiments to demonstrate the performance of the CCE asymmetric learning scheme in unsupervised feature extraction and different downstream tasks with real-world datasets. The efficacy of the proposed Nystr\"om method is also verified to \fra{efficiently} compute the \ksvd.
\end{itemize}

\begin{figure}
    \centering
    \includegraphics[scale=0.46]{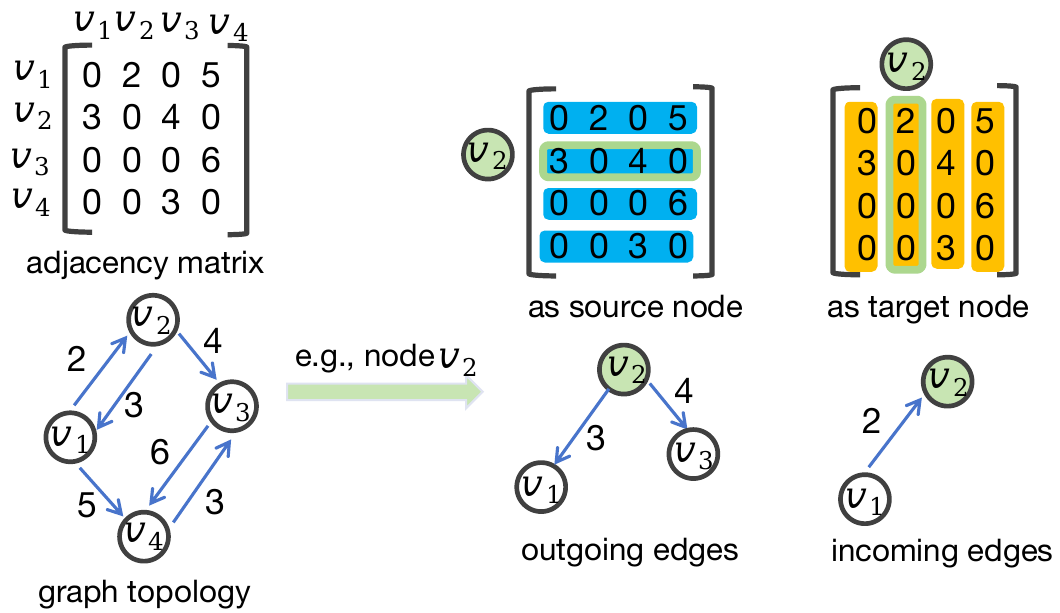}
    \caption{Illustrative example of asymmetric similarity. In a directed graph,
    each node can act as the source or the target. Given the {adjacency matrix $[a(v_i, v_j)]_{i,j=1}^N$}, its rows relate to the outgoing edges, while the columns relate to the incoming edges. The connections between nodes are directional, s.t.~$a(v_i, v_j) \neq a(v_j, v_i), \, i\neq j$. 
    }
    \label{fig:high:level}

\end{figure}

Note that we do not claim to propose the \ksvd algorithm, as it was already sketched in the letter by \citet{suykens2016svd}.
Rather, we give a novel asymmetric learning problem based on two covariance operators in the feature space, whose solution coincides with a \ksvd with infinite-dimensional feature maps, a case that was not previously possible.

%% file: 3-Background.tex
\section{Learning in Feature Spaces with Asymmetry}
\label{sec:learning_asymmetry}
We begin this section by reviewing in Section~\ref{sec:asymmetry} the concept of asymmetric similarity that is critical to this work, before introducing in Section~\ref{sec:cce} the Coupled Covariances Eigenproblem (CCE) that allows us to learn in feature spaces with asymmetry as the solution is ultimately obtained from the SVD of an asymmetric similarity matrix.
We conclude in Section~\ref{sec:related_work_ksvd} with some remarks about related work.
\subsection{Asymmetric Similarity}
\label{sec:asymmetry}
Typically, a kernel $\hat{\kappa} \colon \mathcal{X} \times \mathcal{X} \to \mathbb{R}$ is induced by a \textit{single} feature map $\hat{\phi}$ on a \textit{single} data set whose samples lie in a space $\mathcal{X}$ and is symmetric.
However, in practice, asymmetric similarities are widely used such as in directed graphs (where similarity is directional) as exemplified in Fig.~\ref{fig:high:level}.
Each node acts as source and target and is associated with two feature vectors $x_i,z_i$, possibly from different spaces, for its source and target role, respectively. One can thus extract two sets of features for each node, one related to the nodes to which it points and one for the nodes that point to it.
In general, an asymmetric kernel $\kappa \colon \mathcal{X} \times \mathcal{Z} \to \mathbb{R}$ describes a similarity between elements from \textit{two different} spaces $\mathcal{X}, \mathcal{Z}$.
Despite the utility of asymmetry, classical Mercer-kernel methods, \eg KPCA, only deal with symmetric similarities induced by a single feature map, and thus one has to resort to symmetrizing an asymmetric similarity matrix $K$, which can be done by considering \fra{$({K}^\top+{K})/2$, ${K} {K}^\top$, or ${K}^\top {K}$}.

\input{5a-Related}

%% file: 5a-Related.tex
Compared to the literature on Mercer kernels, asymmetric kernels are less studied.
They have been mostly applied
in supervised learning, e.g., regression \cite{mackenzie2004asymmetric,wu2010asymmetric,kuruwita2010density} and classification \cite{munoz2003support,koide2006asymmetric,2006esann}. 
Some works do not resort to symmetrization: \cite{he2022learning} applies two feature mappings to the given samples and maintains an asymmetric kernel in the LSSVM classifier.
\citet{chen2023primal} applies the variational objective from \citep{suykens2016svd} as an auxiliary regularization loss to the model for low-rank self-attention in Transformers are built as the asymmetric similarity between queries and keys.
Relaxations of the Mercer conditions have also been generalized to learning in reproducing kernel Banach spaces \cite{zhang2009reproducing,xu2019generalized} and Kre\u{\i}n spaces \cite{pmlr-v80-oglic18a}.
Other related but orthogonal approaches include \cite{neto2023kernel} for robust SVD estimation with Gaussian norm in the original space, and \citep{vasilescu2009multilinear} for tensor data where SVD is applied to the symmetric kernel in each mode.

%% file: 4-Method.tex
\subsection{Coupled Covariances Eigenproblem}
\label{sec:cce}
\begin{figure}[t!]
    \centering
    \includegraphics[width=0.6\linewidth]{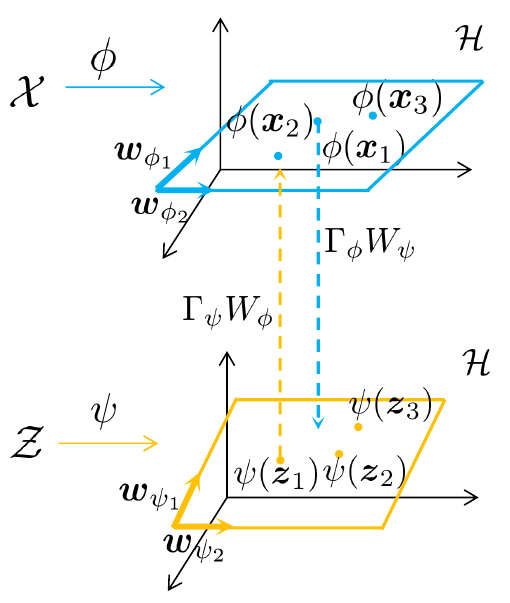}
    \caption{Schematic of our construction. $\mathcal{X},\mathcal{Z}$ from $A$ are mapped to a possibly infinite-dimensional space $\mathcal{H}$. We propose to consider coupled scalar products $\psi(z_j)$ onto $w^\phi_l$ and $\phi(x_i)$ onto $w^\psi_l$. $\mathcal{H}$ is shown separately for clarity.}
    \label{subfig:pdf}
\end{figure}
\begin{figure}[t!]
    \begin{minipage}[t]{.48\linewidth}
        \centering \textbf{KPCA}
        \begin{align*}
            \Sigma_\phi \teal{w_\phi}&=\lambda_\phi \teal{w_\phi}\\
            \Sigma_\psi \brown{w_\psi}&=\lambda_\psi \brown{w_\psi}
        \end{align*}
    \end{minipage}\hfill
    \begin{minipage}[t]{.48\linewidth}
        \centering \textbf{CCE}
        \begin{align*}
            \Sigma_\phi \brown{w_\psi}&=\lambda \teal{w_\phi}\\
            \Sigma_\psi \teal{w_\phi}&=\lambda \brown{w_\psi}
        \end{align*}
    \end{minipage}
    \caption{Overview comparison of KPCA and CCE.}\label{fig:kpca:cce}
\end{figure}
The goal of this section is to gradually define and solve the Coupled Covariances Eigenproblem (CCE).
Our goal is to provide a new tool able to learn in (infinite-dimensional) feature spaces and take advantage of asymmetry.
\paragraph{Notation.}
Given a bounded linear operator $\Gamma$ between Hilbert spaces, its adjoint is referred to as $\Gamma^*$. 
The Frobenius norm of a matrix is denoted by $\norm{\cdot}_\fro$.
The identity matrix of size $r$ is $I_r$.
Set the spaces $\mathcal{X} = \mathbb{R}^m$ and $\mathcal{Z} = \mathbb{R}^n$.
We assume access to two sets of samples $\{x_i\}_{i=1}^n \in \cX^n$ and $\{z_j\}_{j=1}^m \in \cZ^m$.
We consider two mappings $\phi \colon \mathcal{X} \to \mathcal{H}$ and $\psi \colon \mathcal{Z} \to \mathcal{H}$ whose outputs lie in a common feature space $\mathcal{H}$.
We moreover assume that the feature maps are centered. In practice, given the training samples, one can realize the centering by the translated feature maps  $\tilde{\phi}(x) = \phi(x) - \frac{1}{n} \sum_{i=1}^n \phi(x_i)$ and $\tilde{\psi}(z) = \psi(z) - \frac{1}{m} \sum_{j=1}^m \psi(z_i)$, and then the similarity matrix of interest $[\tilde{G}]_{ij} = \langle \tilde{\phi}(x_i) , \tilde{\psi}(z_j)\rangle$ can be computed straightforwardly, e.g., \fra{$\tilde{G} = (I_n-\frac{1}{n} \mathbf{1}_n \mathbf{1}_n^\top)G(I_m-\frac{1}{m} \mathbf{1}_m \mathbf{1}_m^\top)$}.
\par
\paragraph{Construction of the Subspaces in $\cH$.}
In CCE, the goal is to learn a pair of $r$ directions in the feature space $\cH$ that solve a coupled eigenvalues problem.
The sough-after directions are collected in vectors $W_\phi \in \cH^{r}, W_\psi \in \cH^{r}$ as follows:
\begin{align*}
    W_\phi = [w^\phi_{1}, \dots, w^\phi_{r}], &&
    W_\psi = [w^\psi_1, \dots, w^\psi_r].
\end{align*}
Denote by $\Sigma_\phi, \Sigma_\psi \in \cL(\cH)$ the empirical covariance operators described by
\begin{align*}
    \Sigma_\phi = \frac{1}{n} \sum_{i=1}^n \phi(x_i) \phi(x_i)^*, &&
    \Sigma_\psi = \frac{1}{m} \sum_{j=1}^m \psi(z_j) \psi(z_j)^*.
\end{align*}
While performing KPCA would result in solving two eigenvalue problems independently for both covariance operators and using the top $r$ eigenvectors of each to compute interesting directions, we propose to intricate the learned directions in the feature space by solving the following CCE problem:
\begin{definition}[CCE]
    \label{def:cce_problem}
    Find $W_\phi \in \cH^{r}, W_\psi \in \cH^{r}$ such that
    \begin{align}
        \label{eq:cce}
        \Sigma_\phi {W_\psi} = \Lambda W_\phi, &&
        \Sigma_\psi {W_\phi} = \Lambda W_\psi,
    \end{align}
    for some diagonal matrix $\Lambda \in \R^{r \times r}$ with positive values.
\end{definition}
Even if $\cH$ is infinite-dimensional, we can parameterize the directions $W_\phi, W_\phi$ using matrices.
Indeed, given that a solution exists, it holds that for any $l \in \{1, \ldots, r\}$
\begin{equation*}
    \Sigma_\phi w^{\psi}_{l} = \frac{1}{n} \sum_{i=1}^n \langle \phi(x_i), w^\psi_{l} \rangle \phi(x_i) = \lambda_l w^\phi_{l}.
\end{equation*}
Thus all directions $\{w^\phi_l\}_{l=1}^r$ lie in $\Span{\{\phi(x_i)\}_{i=1}^n}$.
Consequently, we can parameterize the directions $W_\phi$ over the $\{\phi(x_i)\}_{i=1}^n$ by a matrix of coefficients $B_\phi \in \R^{n \times r}$. A similar argument holds for the directions $W_\psi$ over the $\{\psi(z_j)\}_{j=1}^m$ with coefficients $B_{\psi} \in \R^{m \times r}$ so that for all $l \in \{1, \ldots, r\}$
\begin{align}
    \label{eq:relation_feature_directions}
    w^\phi_{l} = \sum_{i=1}^n b^\phi_{il} \phi(x_i), &&
    w^\psi_l = \sum_{j=1}^m b^\psi_{jl} \psi(z_j).
\end{align}
\paragraph{Projection Operators.}
Let $\Gamma_\phi \colon \cH^r \to \R^{n \times r}$ and $\Gamma_\psi \colon \cH^r \to \R^{m \times r}$ be linear operators acting on some directions $W \in \cH^r$ in the following way:
\begin{align*}
    [\Gamma_\phi W]_{il} = \frac{1}{\sqrt{n}}\langle \phi(x_i), w_l \rangle, \quad
    [\Gamma_\psi W]_{jl} = \frac{1}{\sqrt{m}}\langle \psi(z_j), w_l \rangle.
\end{align*}
These operators compute the inner products between the chosen directions and the feature maps associated with the data.
As $\Gamma_\phi$ and $\Gamma_\psi$ are bounded linear operators they admit adjoint operators whose action can be made explicit: for any $B \in \R^{n \times r}$, $\Gamma_\phi^* B = \frac{1}{\sqrt{n}}[\sum_{i=1}^n b_{il} \phi(x_i)]_{l=1}^r \in \cH^r$ and $\Gamma_\psi^*$ can be treated similarly.
This observation allows us to rewrite Equation~\ref{eq:relation_feature_directions} under the form
\begin{align*}
    W_\phi = \Gamma_\phi^* B_\phi, && W_\psi = \Gamma_\psi^* B_\psi.
\end{align*}
We also remark that the covariance operators $\Sigma_\phi$ and $\Sigma_\psi$ can be expressed using these projection operators, so that Equation~\ref{eq:cce} can be reformulated using matrix variables $B_\phi, B_\psi$ as
\begin{align}
    \label{eq:cce_gamma}
    \Gamma_\phi^* \Gamma_\phi \Gamma_\psi^* B_\psi = \Gamma_\phi^* B_\phi \Lambda, &&
    \Gamma_\psi^* \Gamma_\psi \Gamma_\phi^* B_\phi = \Gamma_\psi^* B_\psi \Lambda.
\end{align}
\paragraph*{Asymmetric Kernel Matrix.}
The operators $\Gamma_\psi \Gamma_\phi^*$ and $\Gamma_\phi \Gamma_\psi^*$ are of particular interest and their action can be described by related matrices as formalized in the following.
\begin{proposition}
    \label{prop:adjoint_matrix_G}
    Let $G \in \R^{n \times m}$ such that $g_{ij} = \frac{1}{\sqrt{nm}}\langle \phi(x_i), \psi(z_j) \rangle$.
    For all $B_\phi \in \R^{n \times r}$ and $B_\psi \in \R^{m \times r}$,
    it holds that
    \begin{align*}
        \Gamma_\psi \Gamma_\phi^* B_\phi = G^\top B_\phi, && \Gamma_\phi \Gamma_\psi^* B_\psi = G B_\psi.
    \end{align*}
\end{proposition}
This proposition resembles the celebrated \emph{kernel trick} but induces an asymmetry in what is an equivalent of the Gram matrix associated with an asymmetric kernel $\kappa(x, z) = \langle \phi(x), \psi(z) \rangle$.
This kernel permits to avoid the explicit computation of the feature mappings.

Because most classical kernel functions require that the two inputs have compatible dimensions, there are a few challenges associated with the computation of $\kappa(x,z)$ when $\cX$ and $\cZ$ are different by nature.
In this case, we can transform the two inputs $x, z$ into the same dimension through a compatible linear transformation $C \in \cL(\cZ, \cX)$. For Euclidean spaces we can find matrices $C$, such that $C^\top x$ is compatible with $z$ in dimensions, and then
apply existing (symmetric) kernel functions thereafter.

\begin{remark} [Dimensionality Compatibility Matrix]\label{rmk:c:matrix}
 We consider different alternatives to attain the compatibility matrix $C$ as follows: 
$a_0$) the pseudo-inverse of the tackled data matrix; however, it can be computationally unstable and expensive, thus we propose the following $a_1$-$a_3$.

$a_1$) PCA projection on $x_i$; it finds the projection directions capturing the most variance of data samples \citep{pca}.
$a_2$)  randomizing the projection  $C$;  the random linear transformation has been shown to retain the main patterns of the data matrix \citep{larsen2017optimality}.
{$a_3$) learnable $C$ w.r.t.~the downstream tasks; it gives the optimal $C$ by optimizing the downstream task objective, e.g.,  classification loss.}

 {$a_2$ is very computationally efficient while learning the optimal $C$ in $a_3$ can take more computation, up to the task and its optimizer, e.g. SGD optimizer with backpropagated $C$ as experimented in Section \ref{sec:general:data}.
 Note that $a_0$-$a_2$ can be applied under general unsupervised setups for feature learning, while $a_3$ is commonly used when considering end-to-end training for the downstream tasks under supervised setups.}

\end{remark}

\paragraph*{Solution to the CCE.}
Solving the CCE gives rise to a generalized shifted eigenvalue problem, as shown in the following proposition.
\begin{proposition} 
    \label{prop:dual}
    Let $G \in \mathbb{R}^{n \times m}$ be the asymmetric kernel matrix from Proposition~\ref{prop:adjoint_matrix_G}.
    The directions $(W_\phi, W_\psi) \in \cH^r$ respectively parameterized by the matrices $(B_\phi, B_\psi) \in \R^{n \times r} \times \R^{m \times r}$ are solution to the CCE problem if and only if $(B_\phi, B_\psi)$ are solution to the generalized shifted eigenvalue problem
\begin{equation} 
    \label{eq:dual}
    \begin{aligned}
        G^\top G B_\psi &= G^\top B_\phi \Lambda,\\
        G G^\top B_\phi &= G B_\psi \Lambda,
    \end{aligned}
\end{equation}
where ${\Lambda} \in \mathbb{R}^{r \times r}$ is a positive diagonal matrix.
\end{proposition}
According to Lanczos’ decomposition theorem \cite{lanczos1958linear}, Problem~\ref{eq:dual} can be solved by taking for $B_{\phi}, B_{\psi}$ the top-$r$ left and right singular vectors of the matrix $G$.
\begin{proposition} 
    \label{prop:cov}
    Let $B_\phi^{\text{svd}}$ (resp.~ $B_\psi^{\text{svd}}$) be top-$r$ left (resp.~right) singular vectors of $G$.
    Then
    \begin{align*}
        W_\phi= \Gamma_\phi^* B_\phi^{\text{svd}}, && W_\psi= \Gamma_\psi^* B_\psi^{\text{svd}}
    \end{align*}
    is a solution to the CCE.
\end{proposition}
We have shown that solving the CCE reduces to an \ksvd problem, with an asymmetric similarity matrix that involves both feature maps.
Once the directions are learned, if we are given some new data $x \in \cX$ or $z \in \cZ$ we can compute the projected feature scores
\begin{align*}
    [\langle \phi(x) , w^\psi_l\rangle]_{l=1}^{\fra{r}}, && [\langle \psi(z) , w^\phi_l \rangle]_{l=1}^{\fra{r}},
\end{align*}
and use these in downstream tasks.

\paragraph{CCE versus 2KPCA.}
The proposed CCE problem gives a new understanding of the set of directions of interest in the feature space, namely $W_\phi$ and $W_\psi$ from Proposition \ref{prop:cov}, arising from the SVD of the asymmetric kernel matrix $G$ and the feature maps $\phi, \psi$.
We note that these directions can also be interpreted as the principal directions associated to the covariance operators of two symmetrized kernels in two separate KPCA problems arising from feature maps $x \mapsto \Sigma_\psi^{1/2} \phi(x)$ and $z \mapsto \Sigma_\phi^{1/2} \psi(z)$, respectively.
In the dual, this corresponds to taking the SVDs of $G G^\top$ and $G^\top G$, which is equivalent to taking the SVD of $G$. We refer to this interpretation as 2KPCA. 
From a computational standpoint, performing 2KPCA or CCE yields the same singular vectors. However, they are significantly different in the modelling from the following perspectives. 
\begin{itemize}
\item 
In 2KPCA, one takes the principal components associated to kernels built via complicated entanglement of $\phi$ and $\psi$. In CCE, the empirical covariances associated to both feature maps appear free from any other factor. 

\item 
The coupling between the two input variables within the feature maps of 2KPCA is realized through the square root of the other covariance, while in CCE the coupling of the input variables naturally arises by crossing the learned directions in \Cref{def:cce_problem}.

\item
For principal component extraction, we need to compute the projections on the singular vectors $W_\phi$ and $W_\psi$ in $\mathcal{H}$, which are essential in downstream tasks to extract the principal components of test points. This can be easily accomplished in  CCE  with explicit directions, while it is not as clear in 2KPCA. 

\end{itemize}

\subsection{Related work}
\label{sec:related_work_ksvd}
We now discuss research areas that are tangent to our topic: asymmetric kernel SVD (\ksvd) and symmetric kernel approaches such as KPCA or KCCA.
\paragraph{Asymmetric Kernel SVD.}
Given a data matrix $A \in \mathbb{R}^{n \times m}$, \citep{suykens2016svd} regards it w.r.t. either the collection of rows $\{A[i,:] \triangleq  x_i \in \mathcal{X}\}_{i=1}^n$ or the collection of columns $\{A[:,j] \triangleq  z_j \in \mathcal{Z} \}_{j=1}^m$.
In the example in Fig.~\ref{fig:high:level}, $\mathcal{X}$ denotes the outgoing edges of source nodes, while $\mathcal{Z}$ denotes the incoming edges of the target. 
\citep{suykens2016svd} proposes a variational principle for SVD with two linear mappings $\phi(x_i)=C_1^\top   x_i, \psi(z_j)=C_2  z_j$ with compatibility transformations $C_1, C_2$ on the rows and columns of $A$. Provided that the compatibility condition $AC_1C_2A=A$ holds, the stationary solutions correspond to the SVD of $A$.
The two mappings can be extended to construct the $n \times m$ matrix $G_{ij}=k(\phi(x_i), \psi(z_j))$, where $k$ is a kernel function allowing to be nonlinear and asymmetric. 
Stationary solutions are then linked to the SVD of $G$ when the regularization hyperparameters are fixed as the top singular values of $G$.
The \ksvd algorithm therefore finds singular vectors of features non-linearly related to the input variables through the SVD of the non-symmetric rectangular matrix $K$.

\paragraph{Symmetric Kernel Approaches with Covariances.}
\label{rmk:kcca}
Our new construction makes it easier to compare \ksvd with other common algorithms based on finding the best approximation of some covariance quantity, which instead work with symmetric kernels in contrast to our work. KPCA applies a nonlinear feature mapping $\teal{\phi}$ to a set of data samples $  x_i$ and considers projections $\teal{a_{\phi_1}}^{\top} \teal{\phi}(x_i)$ for maximal variances w.r.t.  a \emph{single} covariance ${{\rm cov}\left(\Phi a_{\phi_1}, \Phi a_{\phi_1} \right)}$. 
KPCA can also be tackled through a symmetric PSD kernel $k_{\phi}:=\teal{\phi}^\top(\cdot) \teal{\phi}(\cdot)$ \citep{scholkopf1998}, while \ksvd works with two covariances coupled by each other. We note that doing two KPCA with $\teal{\phi}(x_i)$ and $\red{\psi}(z_j)$ lead to two decoupled covariances and lead to two symmetric kernels $\teal{\phi}^\top(\cdot) \teal{\phi}(\cdot)$ and $\red{\psi}^\top(\cdot) \red{\psi}(\cdot)$ w.r.t. $x_i$ and $z_j$, respectively. This  is significantly different from \ksvd, as shown in Figure \ref{fig:kpca:cce},
as \ksvd is associated with two coupled covariances and essentially works with an asymmetric kernel $\teal{\phi}^\top(\cdot) \red{\psi}(\cdot)$.
KCCA deals with samples from two data sources and only considers projections of each data source.
\ksvd seeks maximal variances of two sets of projections from a single matrix.
Specifically, KCCA considers projections $\teal{a_{\phi_1}}^{\top} \teal{\phi}(x_i)$ and $\red{a_{\psi_1}}^{\top} \red{\psi}(z_i)$ and couples them in a \emph{single} covariance ${{\rm cov}\left(\Phi a_{\phi_1}, \Psi a_{\psi_1} \right)}$.
In our formulation, we consider $\teal{a_{\phi_1}}^{\top} \red{\psi}(z_j)$ and $\red{a_{\psi_1}}^{\top} \teal{\phi}(x_i)$ leading to \emph{two} covariances
$\text{cov}(\Psi a_{\phi_1}, \Psi a_{\phi_1}), \, \text{cov}(\Phi a_{\psi_1}, \Phi a_{\psi_1})$.
KCCA leads to \emph{two separate symmetric PSD kernels} $k_\phi:=\teal{\phi}^\top(\cdot) \teal{\phi}(\cdot)$, $k_\psi:=\red{\psi}^\top(\cdot) \red{\psi}(\cdot)$, while \ksvd couples two feature mappings inducing a \emph{single asymmetric kernel} $\kappa:=\teal{\phi}^\top(\cdot) \red{\psi}(\cdot)$.
Our construction is therefore key to allow for asymmetric kernels w.r.t. KCCA and contrasts with earlier \ksvd constructions, where drawing parallels with related approaches such as KCCA was notably challenging due to the lack of a covariance and subspace interpretation.

%% file: 5-Nystrom.tex
\section{Nystr\"om Method for Asymmetric Kernels}
\label{sec:nystrom}
We adapt the celebrated \nystrom method to asymmetric kernels, the goal being to speed up the computation of the left and right singular vectors of $G$ from Section~\ref{sec:learning_asymmetry}.
The existing \nystrom method approximates eigenfunctions of the integral operator associated with a symmetric kernel \citep{NIPS2000_19de10ad}.
\citet{schmidt1907theorie} discusses the treatment of the integral equations with an asymmetric kernel for the continuous analog of SVD  \citep{stewart1993early}. 
\qh{In this section, we base our formulation upon the pair of adjoint eigenfunctions originally studied in  \citep{schmidt1907theorie}, \qh{namely singular functions}, and start from the corresponding integral equations \citep{bakerprenter1981numerical} to formally derive the asymmetric \nystrom method in a similar spirit with the widely adopted symmetric \nystrom method \citep{NIPS2000_19de10ad}.}

\paragraph{Adjoint Eigenfunctions}   
With an asymmetric kernel $\kappa(   x,    z)$,  $u_s(   x)$ and $v_s(   z)$ satisfying 
\begin{equation}\label{eq:integral:svd}
\begin{array}{l}
     \lambda_s  u_s(   x) = \int_{\mathcal{D}_z} \kappa(   x,    z) v_s (   z) \, p_z(   z) d   z,  \\
    \lambda_s v_s(   z) = \int_{\mathcal{D}_x}  \kappa(   x,    z)u_s (   x) \, p_x(   x) d   x
\end{array}
\end{equation}
are called a pair of adjoint eigenfunctions corresponding to the eigenvalue $\lambda_s$ with $\lambda_1 \geq\lambda_2 \geq \ldots \geq 0$, where $p_x(   x)$ and $p_z(    z)$ are the probability densities over $\mathcal{D}_x$ and $\mathcal{D}_z$. 
Note that \citep{schmidt1907theorie} works with the reciprocal of $\lambda_s$, which is called a singular value by differentiating from the eigenvalues of symmetric matrices \citep{stewart1993early}. 
The integral equations in (\ref{eq:integral:svd}) do not specify the normalization of the adjoint eigenfunctions, which correspond to the left and right singular vectors with finite sample approximation, while generally in SVD  the singular values are solved as orthonormal. Thus, to correspond the results of the adjoint eigenfunctions to the orthonormal singular vectors in SVD, the scalings determining the norms are implicitly included in (\ref{eq:integral:svd}).  
For normalization, {we incorporate three scalings  $l_{\lambda_s}, l_{u_s}, l_{v_s}$ for $\lambda_s, u_s(   x)$, $v_s(   z)$,  respectively, into  (\ref{eq:integral:svd}), such that    $ l_{\lambda_s}\lambda_s  l_{u_s} u_s(   x) = \int_{\mathcal{D}_z} \kappa(   x,    z)  l_{v_s} v_s (   z) \, p_z(   z) d   z$ and $
     l_{\lambda_s}\lambda_s  l_{v_s} v_s(   z) = \int_{\mathcal{D}_x}  \kappa(   x,    z) l_{u_s} u_s (   x) \, p_x(   x) d   x$.}

\paragraph{Nystr\"om Approximation for the Adjoint Eigenfunctions} Given the i.i.d.~samples  $\{   x_1, \ldots,    x_n \}$ and $\{   z_1, \ldots,    z_m \}$, similar to \citep{NIPS2000_19de10ad},
from the probability densities   $p_x(   x), p_z(   z)$ over $\mathcal D_x, \mathcal{D}_z$, 
the  two integral equations in (\ref{eq:integral:svd}) over $p_x(   x)$ and $p_{z}(   z)$ are approximated by   an empirical average:
 \begin{equation}\label{eq:integral:svd:approx}
 \begin{array}{l}
   \lambda_s u_s(   x)  \approx \dfrac{l_{v_s}} {m  l_{\lambda_s}  l_{u_s} }\sum\limits_{j=1}^m \kappa(   x,    z_j)  v_s (   z_j), \\
  \lambda_s  v_s(   z)  \approx   \dfrac{l_{u_s}} {n l_{\lambda_s}   l_{v_s}}\sum\limits_{i=1}^n \kappa(   x_i,    z) u_s (   x_i),
\end{array}
\end{equation}
where $s=1, \ldots, r$, which corresponds to the rank-$r$ compact SVD on a kernel through the Lanczos’ decomposition theorem \cite{lanczos1958linear}:
\begin{equation}\label{eq:integral:svd:motivate}
\begin{array}{l}
       G^{(n, m)} V^{(n, m)}  = U^{(n, m)}\Lambda^{(n, m)}, \\
        (G^{(n, m)})^\top {U}^{(n, m)} =  {V}^{(n, m)}\Lambda^{(n, m)},
\end{array}
\end{equation}
where $G^{(n, m)} \in \mathbb R^{n\times m}$  is the asymmetric kernel matrix with  entries $G_{ij} = \kappa(   x_i,    z_j)$ and $r \leq \min \{n, m\}$, $V^{(n, m)} = [   v^{(n, m)}_1, \ldots,    v^{(n, m)}_r] \in \mathbb R^{m\times r}, U^{(n, m)} = [   u^{(n, m)}_1, \ldots,    u^{(n, m)}_r] \in \mathbb R^{n \times r}$ are column-wise orthonormal and contain the singular vectors,  and $\Lambda^{(n, m)} = {\rm{diag}}\{\lambda_1^{(n, m)}, \ldots,\lambda_r^{(n, m)}\}$ denotes the positive singular values.
To match (\ref{eq:integral:svd:approx})  against (\ref{eq:integral:svd:motivate}), we first require the scalings on the right side of the two equations  in  (\ref{eq:integral:svd:approx}) to be consistent, i.e.,
${l_{v_s}} / ({m  l_{\lambda_s} l_{u_s}}) \triangleq {l_{u_s}} / ({n l_{\lambda_s}  l_{v_s}})$, which yields  
$
   l_{v_s} = \left (\sqrt{m} / \sqrt{n}  \right) l_{u_s}$ and $ {l_{v_s}} / ({m  l_{\lambda_s} l_{u_s}}) \triangleq {l_{u_s}} / ({n l_{\lambda_s}  l_{v_s}}) = {1}/ \left (\sqrt{mn}l_{\lambda_s}\right ).
$

When running all samplings $   x_i$ and $   z_j$  in (\ref{eq:integral:svd:approx}) to match  (\ref{eq:integral:svd:motivate}), 
we arrive at:
$
    u_s(   x_i) \approx \sqrt{\sqrt{mn}l_{\lambda_s}} U^{(n, m)}_{is}$,  $v_s(   z_j) \approx \sqrt{\sqrt{mn}l_{\lambda_s}} V^{(n, m)}_{j s}$, $ \lambda_s \approx ({1} /({\sqrt{mn}}l_{\lambda_s}))\lambda^{(n, m)}_s.
$
The Nystr\"om approximation to the $s$-th pair of adjoint eigenfunctions with an asymmetric kernel  $\kappa(   x,    z)$ is  obtained for $s=1, \ldots, r$:
\begin{equation} \label{eq:nystrom:asymmetric}
\begin{aligned}
    u_s^{(n, m)}(   x) & \approx ({\sqrt{\sqrt{mn}l_{\lambda_s}}}/ \lambda_s^{(n, m)} ) \sum\nolimits_{j=1}^m \kappa(   x,    z_j) V^{(n, m)}_{j s}, \\
  v^{(n, m)}_s(   z) & \approx ({\sqrt{\sqrt{mn}l_{\lambda_s}}} /\lambda_s^{(n, m)}) \sum\nolimits_{i=1}^n \kappa(   x_i,    z) U^{(n, m)}_{i s}, 
\end{aligned}
\end{equation}
which are also called the out-of-sample extension to evaluate new samples, where the norms of $ u_s^{(n, m)},  v_s^{(n, m)}$ are up to the scaling $l_{\lambda_s}$. In (\ref{eq:nystrom:asymmetric}), it explicitly formalizes the approximated adjoint functions (left and right singular vectors) with the asymmetric kernel $\kappa$ ($G$).

\paragraph{Nystr\"om Approximation to Asymmetric Kernel Matrices} With the asymmetric Nystr\"om approximation derived in  (\ref{eq:nystrom:asymmetric}),  we can apply CCE to a subset of the data with sample size $n <N $ and $m< M$  to approximate the adjoint eigenfunctions at all samplings  $\{   x_i\}_{i=1}^N$ and  $\{   z_j\}_{j=1}^M$. 
We assume the kernel matrix to approximate from \ksvd is $G \in \mathbb R^{N\times M}$ and denote $\tilde{\lambda}_s^{(N, M)},  \tilde{   u}_s^{(N, M)}$, and $\tilde{   v}_s^{(N, M)}$ as the  Nystr\"om approximation of the singular values, and left and right singular vectors of  $G$, respectively.
We then utilize  the Nystr\"om method to approximate the singular vectors of $G$ through  the out-of-sample extension (\ref{eq:nystrom:asymmetric}):
\begin{equation}\label{eq:svd:approx:eigne:sub}
\begin{aligned}
     {\tilde{   u}}_s^{(N, M)} = ({\sqrt{{\sqrt{mn}}l_{\lambda_s}}}/ \lambda_s^{(n, m)} )  G_{N,m}    v_{s}^{(n, m)}, \\
     {\tilde{   v}}_s^{(N, M)} = ({\sqrt{{\sqrt{mn}}l_{\lambda_s}}}/ \lambda_s^{(n, m)} )G_{n, M}^\top    u_{s}^{(n, m)}, 
\end{aligned}
\end{equation}
with $\tilde{\lambda}_s^{(N, M)} =  ({1}/{\sqrt{mn}l_{\lambda_s}})\lambda^{(n, m)}_s$ for $s=1, \ldots, r$, where  $   u^{(n, m)}_{s},    v^{(n, m)}_{s}$ are the left and right singular vectors  to the $s$-th nonzero singular value $\lambda_s^{(n, m)}$ of an $n\times m$ sampled submatrix $ G_{n,m} $, $G_{N,m} \in \mathbb R^{N\times m}$ is  the submatrix  by sampling $m$ columns of $G$, and $G_{n, M}\in \mathbb R^{n\times M}$ is   by sampling $n$ rows of $G$. More remarks \qh{on the developed asymmetric \nystrom method} and comparisons to the existing symmetric one are provided in \Cref{sec:supp:sym:nys:discuss}.

%% file: 6-Experiments.tex
\section{Numerical Experiments} \label{sec:exp}

This section aims to give a comprehensive empirical evaluation of SVD in feature spaces with asymmetric kernels in the formulation discussed above.
In existing works, the potential benefits in applications remain largely unexplored w.r.t. advantages of asymmetric kernels.
The following experiments do not claim that asymmetric kernels are always superior to symmetric ones as it can be problem-dependent.
We consider a variety of tasks, including {representation learning in directed graphs}, biclustering, and downstream classification/regression on general data.
A key aspect of our setup is that we can use the solutions $B_\phi, B_\psi$ to express the nonlinear embeddings without explicitly computing the feature mappings $\{\phi(x_i)\}_{i=1}^n$, $\{\psi(z_j\}_{j=1}^m$, which in our derivation, and differently from previous work, are allowed to be infinite-dimensional.
The effectiveness of our new asymmetric \nystrom method is also evaluated.

\subsection{
{Directed Graphs}}\label{sec:test:graph}

\paragraph{{Setups}}

Unsupervised node representation learning extracts embeddings of nodes from graph topology alone.
We consider five benchmark directed graphs \citep{sen2008collective,yanga16}.
\ksvd is compared with its closely related baselines, i.e.,  PCA, SVD, and KPCA, and also with node embedding algorithms DeepWalk \citep{deepwalk}, a well-known random walk-based approach,  HOPE \citep{ou2016asymmetric}, which preserves the asymmetric node roles with two embedding spaces using network centrality measures, \qh{and also  Directed Graph Autoencoders (DiGAE) \cite{kollias2022directed}.}
All compared methods are unsupervised and require only the adjacency matrix; note that this is different from the common setup of graph neural networks \citep{GNN2022wu} that use additional node attributes on top of graph topology and operate in semi-supervised setups.

We evaluate the downstream applications of node classification and graph reconstruction.
With (K)PCA and DeepWalk, we only obtain one set of embeddings. With SVD, \ksvd, HOPE, and \fra{DiGAE} two sets of embeddings are obtained and then concatenated.
As the adjacency matrix is square, there is no compatibility issue.
We compute $\ell_1, \ell_2$ norms (lower is better ({\small $\downarrow$})) for graph reconstruction and Micro- and Macro-F1 scores (higher is better ({\small $\uparrow$})) for node classification using an LSSVM classifier averaged over 10 trials on the extracted 1000 components following \citep{ou2016asymmetric,he2022learning}. 
KPCA employs the RBF kernel and \ksvd employs the asymmetric kernel 
\begin{equation*}
\kappa_{\rm{SNE}}(  {x},  {z}) = \frac{\exp(-\|  {x}-  {z}\|_2^2/\gamma^2)}{\sum_{  {z^{\prime}}\in \mathcal{Z}}\exp(-\|  {x}-  {z^\prime}\|_2^2/\gamma^2)},
\end{equation*}
also known as the SNE kernel \citep{hinton2002stochastic}, which can be seen as an asymmetric extension of RBF
, and conduct 10-fold cross-validation for the kernel parameter in the same range.
Detailed experimental setups are provided in \Cref{app:setups}.

            \begin{table*}[ht!]
      
        \caption{Results on the node embedding downstream tasks with directed graph datasets.} \label{tab:node:classify:reconstruct}
            \small
        \centering
    \resizebox{1\textwidth}{!}{
        \begin{tabular}{lcHcccHccccccccHcccc}
            \toprule
            \multirow{2}{*}{Dataset} &   \multicolumn{10}{c}{Node classification}  &  \multicolumn{9}{c}{ Graph reconstruction} \\
            \cmidrule(lr){2-11}   \cmidrule(lr){12-20}
          & {F1 Score ($\uparrow$)} & {NERD} & PCA& KPCA & SVD &  \ksvd (RBF) & \ksvd  & DeepW& HOPE & DiGAE & Norm ($\downarrow$)& PCA& KPCA & {SVD} & {\ksvd } & {\ksvd} & DeepW& HOPE & DiGAE \\
        
            \midrule
            \multirow{2}{*}{Cora}       & Micro  & 0.733  & 0.757  & 0.771  & 0.776  & 0.787  &  \textbf{0.792}  &  0.741 & 0.750& 0.783 & $\ell_1$ & 556.0 & 349.0 & 622.0 & \textbf{56.0} & \textbf{14.0} & 19.0 & {15.0} &26.0  \\ 
                                        & Macro & 0.724  & 0.751& 0.767 & 0.770  & 0.778   & \textbf{0.784}       &0.736&0.473 & 0.776  & $\ell_2$& 41.2 & 37.9 &41.7 & 19.1 & \textbf{17.4} & {17.6} & 18.1 & 20.9 \\\midrule
            \multirow{2}{*}{Citeseer}   & Micro & 0.507  & 0.648 & 0.666  & 0.667  & 0.675  &  \textbf{0.678} & 0.624 &0.642&0.663 & $\ell_1$ & 138.0 & 46.0 & 176.0 & 45.0 & \textbf{25.0} & \textbf{25.0} & 26.0  & 25.0 \\ 
                                        & Macro & 0.452  &0.611  & 0.635 & 0.632 & 0.637   & \textbf{0.640}  & 0.587& 0.607& 0.627& $\ell_2$& 21.3& 16.0 &24.6 & {14.3} & {14.3} & 14.4 & \textbf{13.3} & 16.4 \\\midrule
            \multirow{2}{*}{Pubmed}    
            & Micro & 0.401  & 0.765
 &0.754  & 0.766&  0.764 &  {0.773} & 0.759 &0.771& \textbf{0.781} & $\ell_1$ & 1937.0 & {171.0} & 1933.0 & {171.0} & \textbf{170.0} & 171.0 & 171.0 &171.0 \\ 
                                        & Macro &  & 0.736 &   0.715  &0.738 &  0.737 & {0.743}   &  0.737  & 0.741& \textbf{0.749}& $\ell_2$ & 128.0 & 31.9 & 118.1& {23.0} & {23.8}  & \textbf{19.4} & 23.8 & 27.9 \\\midrule
                  	\multirow{2}{*}{TwitchPT}     & Micro && 0.681 &	0.681	&0.694	&& \textbf{0.712}  &  0.637  &0.685& 0.633 & $\ell_1$ & 1780.0 &	766.0	& 1839.0 &&	\textbf{756.0} & 864.0 & 1108.0 & 759.0\\
                 & Macro & & 0.517& 0.531	& 0.543	& & \textbf{0.596}   &  0.589  &0.568& 0.593&$\ell_2$  & 196.3 &	172.4 &	192.1	& & \textbf{140.3} & 146.5 & 158.2 & 79.7
                 \\\midrule
               	\multirow{2}{*}{BlogCatalog}     & Micro && 0.648 &	0.663	& 0.687 &	& \textbf{0.710} & 0.688  &0.704& 0.697 &$\ell_1$ & 5173.0 &	766.0 &	5166.0	& & \textbf{764.0} &810.0 & 3709.0 & 771.0\\
                 & Macro & & 0.643 &	0.659 &	0.673	&& \textbf{0.703} &  0.679  &0.697& 0.690& $\ell_2$  & 429.4 &	99.9	& 410.5	&& \textbf{94.2} & 104.0 & 286.7 & 202.6\\
             \bottomrule
        \end{tabular}
        }
    \end{table*}

      \paragraph{{Results}} In \Cref{tab:node:classify:reconstruct} for node downstream classification,
    the results indicate consistent improvements over both SVD and KPCA, verifying the effectiveness of employing nonlinearity (to SVD) and asymmetric kernels (to KPCA). 
    The graph reconstruction task reflects how well the extracted embeddings preserve the node connection structure.
    The adjacency matrix is reconstructed with the learned embeddings and then compared to the ground truth with $\ell_1, \ell_2$ norms.
    Asymmetric kernels greatly improve SVD, further illustrating the significance of using nonlinearity. KPCA achieves better performance than  SVD, 
   showing that considering the asymmetry alone, i.e., SVD, is not enough and nonlinearity is of great importance. 
   Although DeepWalk, HOPE, and DiGAE are designed specifically for graphs, the simpler \ksvd shows competitive performance, demonstrating great potential in representation learning for directed graphs.

\subsection{Biclustering}\label{sec:Biclustering}
\paragraph{{Setups}} Biclustering simultaneously clusters samples and features of the data matrix, e.g., cluster documents and words. 
SVD has long been a common method by clustering rows and columns through right/left singular vectors.  
KPCA can be applied either to the rows or the columns at a time, due to its symmetry. We apply $k$-means to the extracted embeddings from SVD, KPCA, and \ksvd. We also compare 
with the biclustering methods EBC \cite{ebc}, based on ensemble,  and 
the recently proposed BCOT \citep{fettal2022efficient}, based on optimal transport. In the considered benchmarks \citep{fettal2022efficient}, the rows relate to documents, where the NMI metric can be used. The columns relate to terms, where the Coherence index is used
\citep{pmi}.
Other settings are as in \Cref{sec:test:graph} and we use $a_1$ for the compatibility matrix.

\paragraph{Results} In Table \ref{tab:biclustering}, \ksvd outputs considerably better clustering compared to KPCA, which can only perform clustering on a single data view at a time. 
Despite the \ksvd algorithm not being specialized for this task, it consistently achieves competitive or superior performance compared to BCOT and EBC, both specifically designed for biclustering. This experiment further emphasizes the significance of asymmetric feature learning and its potential to boost the performance of downstream tasks in applications.

    \begin{table}[t]
    	\caption{{Biclustering results of documents w.r.t. NMI and of terms w.r.t. Coherence (Coh).}}\label{tab:biclustering}
    	\centering
         \resizebox{\linewidth}{!}{
    	\begin{tabular}{ccccccccc}
    		\toprule
    	   \multirow{2}{*}{Method}	 & \multicolumn{2}{c}{ACM}     &  \multicolumn{2}{c}{DBLP}    &   \multicolumn{2}{c}{Pubmed} &   \multicolumn{2}{c}{Wiki}	\\	
         \cmidrule(lr){2-9} 
         & NMI & Coh & NMI & Coh& NMI & Coh& NMI & Coh \\
    		\midrule
      SVD& 0.58 & 0.21& 0.09 & -0.06& 0.31 & 0.42& 0.39 &0.42  \\
      KPCA& 0.59 &0.28 & 0.26& 0.17 & 0.29& 0.51 & 0.46& 0.57\\
      \ksvd& \textbf{0.68} & \textbf{0.32}& \textbf{0.28} &{0.21} & \textbf{0.33} & 0.54 &\textbf{0.48} & \textbf{0.64} \\
      BCOT& 0.38 &0.27  &0.27& \textbf{0.22}& 0.16 &0.54 & \textbf{0.48}  & \textbf{0.64}\\
      EBC & 0.62 & 0.20& 0.15 & 0.21& 0.19 & \textbf{0.56} & 0.47& 0.63 \\
      \bottomrule
    	\end{tabular}
     }
     
    \end{table}

\subsection{
{General Data}}\label{sec:general:data}
\paragraph{Setups} 
Since asymmetric kernels are more general than symmetric ones, the features learned with asymmetric kernels can help boost performance in generic feature extraction.
We evaluate \ksvd on general data from UCI \citep{Dua:2019}.
First, we extract embeddings with kernel methods, and then apply a linear classifier/regressor and report results on test data (20\% of the dataset).
Besides SNE, we employ RBF and note that the resulting kernel matrix $G$ in \eqref{eq:dual} is still asymmetric, as the kernel is applied to two different sets $\mathcal{X}$ and $\mathcal{Z}$, i.e. $\kappa(   x_i,    z_j)\neq \kappa(   x_j,    z_i)$.
Data matrices are generally non-square, so we need the dimensionality compatibility $C$ as in \Cref{rmk:c:matrix}.
$C$ is realized by $A^\dag$ in previous work \cite{suykens2016svd}; we denote this approach $a_0$.
We compare $a_0$ with our proposed approaches $a_1,a_2$ in unsupervised settings, and with our $a_3$ with learnable $C$, optimized by SGD on the downstream task objective.

    \begin{table*}[ht!]
        \caption{Downstream task results on 
        general UCI datasets, {where the best results are in bold}.} \label{tab:uci}
        \centering
        \resizebox{\textwidth}{!}{
        \small
    \begin{tabular}{cccccccccccc}
        \toprule
        \multirow{2}{*}{Dataset} & \multirow{2}{*}{Metric} & \multirow{2}{*}{\begin{tabular}[c]{@{}c@{}}KPCA\\ (RBF)\end{tabular}} & \multirow{2}{*}{\begin{tabular}[c]{@{}c@{}}KPCA\\ (RBF Learnable $C$)\end{tabular}}  & \multicolumn{4}{c}{\ksvd (RBF)}           
        & \multicolumn{4}{c}{\ksvd (SNE)}           
        \\   \cmidrule(lr){5-8} \cmidrule(lr){9-12}
        & & & & $a_0$ & $a_1$ & $a_2$ & Learnable $C$ ($a_3$) 
        & $a_0$ & $a_1$ & $a_2$ & Learnable $C$ ($a_3$) 
        \\ \midrule
        Diabetes & ACC ($\uparrow$) 
        &  {0.67} & 0.68& 0.66  &  {0.67} &  {0.67}  & {\bf 0.71} 
        & 0.66  & 0.65 &  {0.67}  & {\bf 0.69} 
        \\
        Ionosphere & ACC ($\uparrow$)    
        & 0.67 & 0.68 & {0.68}  &  {0.69} &  {0.69}  & {\bf 0.70} 
        &  {0.70}  & {0.67}  & {0.68}  & {\bf 0.72}  
        \\
        Liver & ACC ($\uparrow$) 
        & 0.71  & 0.72&  {0.74}  & 0.70  & {0.71}  & {\bf 0.76} 
        & {0.71}&  { 0.72} & 0.70  & {\bf 0.75}   
        \\
        Cholesterol & RMSE ($\downarrow$) 
        & 59.36 & 53.14 & {54.53}  &  {52.80}  & {55.24}  & {\bf 45.91} 
        & {49.11}&   {47.83}  & {48.12}  & {\bf 44.40}   
        \\
        Yacht & RMSE ($\downarrow$)                  
        & 15.85& 14.05 & {14.90} &  {14.74}  & 16.57 & {\bf 12.98} 
        & {15.41}&   {14.18} &  {13.89}  & {\bf 13.05}  
        \\
        Physicochemical-protein & RMSE ($\downarrow$)  
        & 5.96 &5.96 & 5.94 & 5.96 & 6.01 & {\bf 5.90} 
        & 5.96 & 6.03 & 6.00 & {\bf 5.93}
        \\
        \bottomrule             
        \end{tabular}}
    \end{table*}

\paragraph{Results} In Table \ref{tab:uci}, 
\ksvd maintains the best overall results with all alternatives $a_0$-$a_3$, showing 
promising potentials of applying asymmetric kernels on general data for downstream tasks.
Under unsupervised setups, the alternatives $a_1$-$a_2$ for $ C$  all lead to comparable performance to the expensive pseudo-inverse $a_0$.
For fair comparisons with learnable $C$, we also evaluate KPCA with optimized $C$, i.e., we use $\hat{\kappa}(C^\top    x, C^\top    x)$ in KPCA.
With $a_3$, asymmetric kernels consistently outperform KPCA, while, for KPCA, a learnable $C$ only provides marginally improved or comparable results.
The matrix $C$  can be viewed as a transformation for dimensionality compatibility providing additional degrees of freedom to learn enhanced embeddings.

  \begin{table}[ht!]
        \caption{	
        Runtime for multiple \ksvd problems at tolerance $\epsilon = 10^{-1}$; the lowest solution time is in bold. 
        }
        \label{tab:speedup:low:high}
        \centering
        \resizebox{\linewidth}{!}{
        \begin{tabular}{lcccccc}
            \toprule
            \multirow{2}{*}{Task} & \multirow{2}{*}{$N$} & \multicolumn{4}{c}{Time (s)} 
            & \multicolumn{1}{c}{} \\\cmidrule(lr){3-7} 
            & & TSVD & RSVD & Sym.~Nys. & Ours & Speedup\\	
            \midrule
            Cora       & 2708  & 0.841  &  0.274  & 0.673   & \textbf{0.160}  & 1.71$\times$  \\ 	
            Citeseer   & 3312  & 0.568  &  0.290  & 0.214   & \textbf{0.136}  & 2.14$\times$ \\ 	
            PubMed     & 19717 & 9.223  &  4.577  & 44.914  & \textbf{0.141}  & 32.51$\times$\\ 
            \bottomrule
        \end{tabular}
        }
    \end{table}

\subsection{Asymmetric Nystr\"om Method}
We evaluate the proposed asymmetric Nystr\"om method against other standard solvers on problems of different sizes.
We compare with three common SVD solvers: truncated SVD (TSVD) from the ARPACK library, the symmetric Nystr\"om (Sym.~Nys.) applied to $GG^\top$ and $G^\top G$ employing the Lanczos Method \citep{lehoucq1998} for the SVD subproblems, and randomized SVD (RSVD) \citep{halko2011}.
For all used solvers, we use the same stopping criterion based on achieving a target tolerance $\varepsilon$.
The accuracy of a solution $\tilde{U}=[\tilde{   u}_1,\dots,\tilde{   u}_r], \tilde{V}=[\tilde{   v}_1,\dots,\tilde{   v}_r], $ is evaluated as the weighted average 
$\eta = \frac{1}{r} \sum_{i=1}^r w_i ( 1 - |   u_i^\top \frac{\tilde{   u}_i}{\norm{\tilde{   u}_i}}| ) +
\frac{1}{r} \sum_{i=1}^r w_i ( 1 - |   v_i^\top \frac{\tilde{   v}_i}{\norm{\tilde{   v}_i}}| )$,
with $w_i=\lambda_i$ and
 $U=[   u_1,\dots,   u_r], V=[   v_1,\dots,   v_r]$  the left and right singular vectors of $G$ from its rank-$r$ truncated SVD. 
The stopping criterion for all methods is thus $\eta \leq \varepsilon$. 
This criterion is meaningful in feature learning tasks as the aim is to learn embeddings of the given data as scalar products with the singular vectors, rather than approximating the full kernel matrix.
We use random subsampling for all Nystr\"{o}m methods and increase the number of subsamples $m$ to achieve the target $\varepsilon$, where we use $m=n$ as the kernel matrices are square; we employ the SNE kernel and set $r=20$.

Table \ref{tab:speedup:low:high} shows the algorithm running time at tolerance level $\varepsilon=10^{-1}$.
We also show the speedup w.r.t.~RSVD, i.e., $t^\text{(RSVD)}/t^\text{(Ours)}$, where $t^\text{(RSVD)}, t^\text{(Ours)}$ denote the training time of RSVD and our asymmetric \nystrom solver.
Our solver shows to be the fastest and our improvement is more significant with larger problem sizes. In Appendix \ref{sec:appendix:results:nystrom}, we present the results at tolerance level $\varepsilon=10^{-2}$, also verifying our advantages.
Further, we consider that a solver's performance may depend on the singular spectrum of the kernel.
We vary the bandwidth $\gamma$ of the SNE kernel on Cora to assess how the singular value decay of the kernel matrix affects performance, where an increased $\gamma$ leads to spectra with faster decay, and vice versa.

\begin{figure}[t!]
\centering
\begin{minipage}{.5\textwidth}
    \centering
    \includegraphics[width=0.95\textwidth]{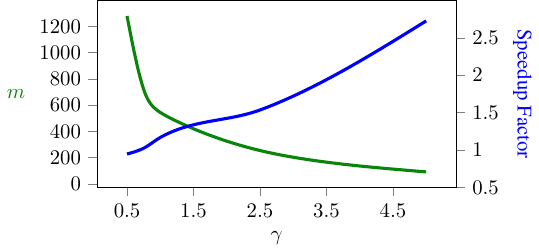}
    \caption{ \textbf{Varying singular spectrum.}
    Number of samples $m$ (green) to achieve a fixed tolerance and the speedup factor w.r.t. RSVD (blue) on Cora when the singular spectrum of $G$ changes (larger $\gamma$ leads to faster decay).
    }
    \label{fig:spectrum}
\end{minipage}%
\hfill
\begin{minipage}{.5\textwidth}
	\centering
    \includegraphics[width=0.7\textwidth]{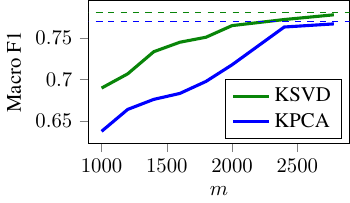}
	\caption{\textbf{Effect of $m$.} Performance on Cora at different $m$ by asymmetric Nystr\"om. Dashed lines indicate the exact solution.}
	\label{fig:nystrom:cora}
\end{minipage}
\end{figure}

In Fig.~\ref{fig:spectrum}, we vary $\gamma$ and show the required subsamples $m$ to achieve the given tolerance and the runtime speedup w.r.t. RSVD.
Our method shows overall speedup to RSVD, and our asymmetric Nystr\"om requires significantly fewer subsamples on matrices with faster singular spectrum decay, showing greater speedup in this scenario.
In Fig.~\ref{fig:nystrom:cora}, the node classification F1 score (Macro) is reported for several values of subsamples $m$, where \ksvd employs the asymmetric Nystr\"om method and KPCA uses the symmetric Nystr\"om on the same RBF kernel. It shows superior performances of the asymmetric method at all considered $m$ without significant accuracy decrease 
due to the subsampling. Additional results are provided in \Cref{app:exp}.

%% file: 7-Conclusion.tex
\section{Conclusion}
\fra{
This work presents a novel learning scheme for asymmetric learning in feature spaces.
We establish that the solution to the coupled covariances eigenproblem (CCE) can be obtained by performing SVD on an asymmetric kernel matrix, providing a new perspective on \ksvd grounded in covariance operators.  
In addition, the resulting computations can be sped up on large-scale problems, thanks to the formally derived asymmetric Nystr\"om method.
Numerical results show the potential of the retained asymmetry and nonlinearity realized in \ksvd and the effectiveness of the developed asymmetric Nystr\"om method.
The insights and methodologies in this work pave the way for further exploration of asymmetric kernel methods in machine learning.  
}

%% file: 9-Appendix.tex
\section{Further comparisons with related work}\label{sec:supp:related:work}

\subsection{\ksvd and  discussions with related work}\label{sec:supp:related:work:ksvd}
Our main interest in this work is to  {derive a new formulation for \ksvd,}
to promote more insights into 
{nonlinear feature learning with considerations to asymmetry.}
We  start from a new asymmetric learning paradigm
based on coupled covariances eigenproblem (CCE) and show that the solutions to CCE  leads to the \ksvd problem associated with a specific asymmetric similarity matrix that
blends in two feature maps. Our formulations involve two covariance operators allowing to work with  infinite-dimensional  feature mappings with induced asymmetric kernels, aiming to provide a vigorous formalization equipped with interpretations w.r.t.  both the covariance matrix and kernel matrix. 
 \ksvd  attains an asymmetric kernel matrix $G$ simultaneously coupling two sets of  mapping information, which is intrinsically different from  KPCA. Through this work, we would also like to convey that although the solutions of PCA and KPCA can be computed numerically by the linear algebra tool of SVD,
PCA is essentially different from SVD, and so is KPCA  from \ksvd.

The solution of \ksvd  leads \eqref{eq:dual} in terms of an asymmetric kernel $G$ instead of the given data matrix $A$,  and is therefore related to the compact SVD as a solution to \eqref{eq:dual}.  
\cite{suykens2016svd} revisits the compact matrix SVD with a variational principle under the setups of least squares support vector machines (LSSVM), where the dual solution leads to a shifted eigenvalue problem regarding the given data matrix $A$.  It  focuses on the (linear) matrix SVD; although it mentions  the possibility with nonlinearity by transforming $A$ into some asymmetric kernel matrix of the same size, it {cannot deal with infinite-dimensional feature spaces
nor connect to the covariances, where 
it neither formalizes the   derivations to the kernel trick, nor mentions possible applications with any experimental evaluations}. {In \cite{chen2023primal},  the asymmetric self-attention is remodelled  for low-rank properties {through the finite-dimensional feature mappings with neural networks}.The queries and keys are regarded as two data sources and directly tackle the self-attention  by applying the variational objective proposed  in \cite{suykens2016svd} as an auxiliary regularization loss into  the optimization objective, which is iteratively minimized  to approach zero and cannot provide the singular vectors nor singular values.} In the early work of  Schmidt \cite{schmidt1907theorie}, the shifted eigenvalue problem  is also discussed w.r.t. the integral equations regarding a pair of adjoint eigenfunctions in the continuous cases with function spaces. Hence, we can see that there can be   multiple frameworks that can lead to a solution in the form resembling a shifted eigenvalue problem either on the given data matrix or an asymmetric kernel matrix as derived in \ksvd, whereas different goals are  pertained in the addressed scenarios and the methodologies are  also varied with different optimization objectives and interpretations. 

Moreover, to get the terminology of \ksvd clearer, we additionally discuss the differences to a few other existing works that share some similarities in naming the methodology. In \cite{neto2023kernel},  it considers a new algorithm for SVD that incrementally estimates each set of robust singular values and vectors by replacing the Euclidean norm with the Gaussian norm in the objective.
Different from kernel-based methods, \cite{neto2023kernel} operates in the original space, not in the feature space, where the kernel is only used in the objective for the estimator and the data are not processed with any nonlinearity in the feature space. Despite the similarity in names, the tasks and methodologies in \cite{neto2023kernel} and \ksvd are intrinsically different. In \cite{he2022learning},  it presents how to apply asymmetric kernels with LSSVMs for supervised classification with both input samples and their labels, \qh{and is derived with finite-dimensional feature spaces.} In  particular, unlike our construction with $\cX$ and $\cZ$, \cite{he2022learning} can only consider  a single data set under the context of its supervised task, exploring the supervised learning for the row data and possibly missing full exploitation of the asymmetry residing in the data. \qh{Accordingly, the asymmetry in \cite{he2022learning} only comes from the choice of the asymmetric kernel function, while our asymmetry also comes from jointly handling two different sets.} In \cite{vasilescu2009multilinear}, KPCA is extended to tensor data to analyze the factors w.r.t. each mode of the tensor, where SVD is applied to solve the eigendecomposition of the KPCA problem in each mode and the left singular vectors (i.e., eigenvectors) are obtained as the nonlinear factor for each mode. \cite{vasilescu2009multilinear} still only considers the symmetry in feature learning but extend it to higher-order tensors. Hence, the data processing, the kernel-based learning scheme, the optimization framework, and also the task are all different from the ones considered in the present work.

\subsection{Asymmetric \nystrom method and related work}\label{sec:supp:related:work:nystrom}

\subsubsection{Background}\label{sec:supp:sym:nys}
The existing  \nystrom method starts from the numerical
treatment of an integral equation with a symmetric kernel  function  $\hat{\kappa}(\cdot, \cdot)$ such that $ \lambda u(x) =  \int_{a}^b \hat{\kappa}( x, z)u ( x) \,d x$, i.e., the continuous analogue to the eigenvalue problem, where the quadrature technique can be applied to formulate the discretized approximation  \cite{bakerprenter1981numerical}. Concerning the more general cases with multivariate inputs, the probability density function and the empirical average technique of finite sampling have been utilized to compute the approximated eigenfunctions that correspond to the eigenvectors \cite{bakerprenter1981numerical,scholkopf1999}. To better illustrate the differences to the established asymmetric \nystrom, we provide more details on the  symmetric \nystrom  method for reference, based on the  derivations from \cite{NIPS2000_19de10ad}.

Given the i.i.d. samples $\{   x_1, \ldots,    x_q \}$ from the probability density 
 $p_x(   x)$ over $\mathcal D_x$, an empirical average is used to approximate the integral of the eigenfunction with a symmetrick kernel: 
\begin{equation}\label{eq:integral:evd}
    \lambda_s u_s(   x) =  \int_{\mathcal{D}_x} \hat{\kappa}(   x,    z)u_s (   x)p_x(   x) \,d   x \approx \frac{1} {q}\sum\nolimits_{i=1}^q \hat{\kappa}(   x,    x_i)u_s (   x_i),
\end{equation}
where $u_s$ is said to be an eigenfunction of $\hat{\kappa}(\cdot, \cdot)$ corresponding to the eigenvalues with $\lambda_1\geq \lambda_2 \geq \ldots \geq 0$. By running $   x$ in \eqref{eq:integral:evd} at $\{   x_1, \ldots,    x_q \}$, an eigenvalue problem is motivated, such that  $G^{(q)}U^{(q)}=U^{(q)}\Lambda^{(q)}$, where $G^{(q)}\in \mathbb R^{q\times q}$ is the Gram matrix with $G_{ij}^{(q)}=\hat{\kappa}(   x_i,    x_j)$ for $i,j =1, \ldots, N$, $U^{(q)}=[   u^{(q)}_1, \ldots,    u^{(q)}_q] \in \mathbb R^{q\times q}$ is column orthonormal and the diagonal matrix $\Lambda^{(q)} \in \mathbb R^{q\times q}$ contains the eigenvalues  such that $\lambda_1^{(q)} \geq \ldots  \geq \lambda_q^{(q)} \geq 0$. In this case, the approximation of eigenvalues and eigenfunction from the integral equation \eqref{eq:integral:evd} arrives at: 
\begin{equation}\label{eq:approx:evd}
 \lambda_s  \approx \frac{\lambda_s^{(q)}}{q}, \quad
     u_s(   x_i) \approx \sqrt{q} U^{(q)}_{i,s},
\end{equation}
which can be plugged back to \eqref{eq:integral:evd}, leading to the \nystrom approximation  to the $i$-th eigenfunction:
\begin{equation}\label{eq:sym:nystrom:u}
    u_s(   x) \approx \dfrac{\sqrt{q}}{\lambda^{(q)}_s}\sum_{i=1}^q {\hat{\kappa}}(   x,    x_i) U^{(q)}_{i,s},
\end{equation}
with $\forall s: \lambda^{(q)}_s >0$.
With the \nystrom technique in \eqref{eq:sym:nystrom:u}, one can use different sampling sets to approximate the integral \eqref{eq:integral:evd}. Thus, given a larger-scale Gram matrix  $G^{(N)}\in \mathbb R^{N\times N}$, for the first $p$ eigenvalues and eigenfunctions, a subset of  training data $q\triangleq n< N$ can be utilized to attain their approximation   at all $N$ points for the kernel matrix $G^{(N)}$ with \eqref{eq:approx:evd}: 
\begin{equation}\label{eq:approx:evd:kernel}
      \tilde{\lambda}^{(N)}_s  \triangleq \dfrac{N}  {n}\lambda_s^{(n)},\quad
     {\tilde{   u}}_s^{(N)} \triangleq \sqrt{\dfrac {n}{N}} \dfrac{1}{\lambda_s^{(n)}} G_{N,n}    u^{(n)}_{s}, 
\end{equation}
where $ \tilde{\lambda}^{(N)}_s $ and $  {\tilde{   u}}_s^{(N)}$ are the \nystrom approximation of the eigenvalues and eigenvectos of $G^{(N)}$. Here $    u^{(n)}_{s}$  are  eigenvectors corresponding to the $s$-th   eigenvalues $\lambda_s^{(n)}$ of an $n\times n$ submatrix $G_{n,n}$ and $G_{N,n}$ is the submatrix by sampling $n$ columns of  $G^{(N)}$.

\subsubsection{Discussions}\label{sec:supp:sym:nys:discuss}
We provide the following remarks elaborating on the existing \nystrom method w.r.t. the  eigenvalue problem for  Mercer kernels and {our extended \nystrom method w.r.t.}  the SVD problem for asymmetric kernels.
\begin{enumerate}
    \item \textbf{Integral equations.} As shown in Section \ref{sec:supp:sym:nys} above, the existing \nystrom method starts from  a single integral equation with a symmetric kernel $\hat{\kappa}(\cdot, \cdot)$, corresponding to an eigenvalue problem in the discretized scenarios \cite{bakerprenter1981numerical,NIPS2000_19de10ad}. {Thus, the  existing \nystrom method is derived only for Mercer kernels with  symmetry constraints on the tackled matrix.} Differently, the proposed asymmetric \nystrom method deals with an asymmetric kernel ${\kappa}(\cdot, \cdot)$ and starts from a pair of adjoint eigenfunctions, which jointly determine an SVD problem in the discretized scenarios \cite{schmidt1907theorie,stewart1993early}.  {In \cite{drineas2005nystrom,nystromsvdnemtsov2016matrix},  the matrix compression is discussed with   \nystrom-like methods to general matrices.} However, the method in \cite{nystromsvdnemtsov2016matrix} is formulated to approximate subparts of the left and right singular vectors, and still applies the symmetric \nystrom method  to heuristically  approximate the asymmetric submatrix twice for the corresponding subparts; \cite{drineas2005nystrom} directly applies the symmetric \nystrom method  and resembles its formulas to approximate left and right singular vectors of general matrices, ignoring the asymmetry constraints. Rather than working with singular vectors, {\cite{xiong2021nystromformer} utilizes the technique in \cite{drineas2005nystrom} to the submatrix blocks to approximate a surrogate attention matrix in Transformers for computation efficiency.} Hence, the analytical framework of the asymmetric  \nystrom method has not been formally formulated yet. In our paper, the explicit rationale of leveraging the \nystrom technique is provided for the asymmetric matrices {through the finite sample approximation to the pair of adjoint eigenfunction, which incorporates the asymmetry constraints on the tackled matrix,} so that from analytical and practical aspects it becomes viable  to directly apply the asymmetric 
    \nystrom method to the cases that pertain the asymmetric nature.

    \item \textbf{Special case with symmetry.} In the derivations on the finite sample approximation, three  scalings $l_{\lambda_s}$, $ l_{u_s}$, and $l_{v_s}$ are introduced to the singular  values $\lambda_s$, right singular vectors $u_s(   x)$, and left singular vectors $v_s(   z)$ in Eq.  (7) in Section 4 in the paper, for the considerations on their norms. Meanwhile, the constant coefficients in the two equations in Eq. (8) in the paper are required to be the same in scalings to proceed the derivations that match the SVD problem. 
    In the existing symmetric  \nystrom method, the scaling issue of the approximated eigenfunction does not appear with $ l_{\lambda_s} \lambda l_{u_s} u(   x) =  \int \hat{\kappa} (   x,    z) l_{u_s} u(   x)p_x(   x) \,d   x$, as the scaling $l_{u_s}$  is cancelled out in the two sides of this equation, i.e., the Eq. \eqref{eq:integral:evd} above. Thus, in \eqref{eq:integral:evd} it implicitly sets the scaling  of the eigenvalue as $l_{\lambda_s}=1$  \cite{NIPS2000_19de10ad}, while  in \eqref{eq:approx:evd:kernel} $l_{\lambda_s}$ is set as $1/ N$ in the application of the \nystrom method to speedup the eigenvalue problem on a larger  Gram matrix $G^{(N)}$.

    Note that, for feature learning, we only need to find the singular vectors in Eq. (10) or (11) in the paper, which are taken as embeddings of the given data for downstream tasks.
    The computation of the singular values can be omitted, so that we can simply implement the scaling through normalization in practice.
    The numerical computation of the approximated kernel matrix is also not necessary for the considered feature learning tasks.
    When considering the special case where the kernel matrix $G$ in \ksvd is square ($ N=M$) and symmetric  ($G = G^\top$), the numbers of samplings to the rows and column are the same ($n=m$),  and the scaling $l_{\lambda_s}$ is set the same, the asymmetric \nystrom method boils down to  the existing \nystrom method.

\item \textbf{Another alternative derivation.}
We consider an asymmetric kernel function $  \kappa(x,y)$, and define the induced kernel operator and its adjoint by
\begin{equation}
\begin{aligned}
(Gg)(x)&=\mathbb{E}_{p_x(x)}[  \kappa(x,Y)g(Y)],\\
(G^*f)(y)&=\mathbb{E}_{p_y(y)}[  \kappa(X,y)f(X)],
\end{aligned}
\end{equation}
for $L^2$-integrable functions $f$ and $g$, where we denote the two datasets in the matrix form by arranging the samples row-wisely in $X$ and $Y$, respectively.
Then, the left and right $s$-th singular functions $u_s(\cdot)$ and $v_s(\cdot)$ of the kernel operator $  \kappa(x,y)$ satisfy
\begin{equation}
\begin{aligned}
(G^*u_s)(y)&=\lambda_s v_s(y),\\
(Gv_s)(x)&=\lambda_s u_s(x).
\end{aligned}
\end{equation}

Given $n$ samples $x_1,\ldots,x_n$ drawn from $p_x(x)$ and $m$ samples $y_1,\ldots,y_m$ drawn from $p_y(y)$, the relations can be approximated as
\begin{equation}
\begin{aligned}
v_s(y)&= \frac{1}{\sigma_s} (G^*u_s)(y) \approx \frac{1}{n\lambda_s} \sum_{i=1}^n   \kappa(x_i,y)u_s(x_i),\\
u_s(x)&= \frac{1}{\sigma_s} (Gv_s)(x) \approx \frac{1}{m\lambda_s} \sum_{j=1}^m   \kappa(x,y_j) v_s(y_j).
\end{aligned}
\end{equation}
As $G= [\kappa(x_i, y_j)] =U\Lambda V^T \in \mathbb R^{n\times m}$, we then scale the kernel matrix by $1/{\sqrt{mn}}$, and the left and right singular vectors by $1/\sqrt{n}$ and $1/\sqrt{m}$, respectively, yielding
the approximated estimates of the pair of adjoint eigenfunctions:
\begin{equation}
\begin{aligned}
v^{(n,m)}_s(y)& \approx \frac{1}{n\lambda_s} \sum_{i=1}^n   \kappa(x_i,y)U_{is}^{(n,m)},
\\
u_s^{(n,m)}(x)& \approx \frac{1}{m\lambda_s} \sum_{j=1}^m   \kappa(x,y_j)V_{js}^{(n,m)},
\end{aligned}
\end{equation}
such that
\begin{equation}
\begin{aligned}
v_s(y)& \approx \frac{1}{n\lambda_s} \sum_{i=1}^n   \kappa(x_i,y)u_s(x_i)\Rightarrow \frac{\sqrt{mn}}{N\lambda_s} \sum_{i=1}^n   \kappa(x_i,y_j)\sqrt{n} U_{is} \approx \sqrt{m} V_{js}  \Rightarrow  \frac{1}{\lambda_s} \sum_{i=1}^n   \kappa(x_i,y_j) U_{is} \approx  V_{js}, 
\\
u_s(x)& \approx \frac{1}{m\lambda_s} \sum_{j=1}^m   \kappa(x,y_j)v_s(y_j)
 \Rightarrow \frac{\sqrt{mn}}{m\lambda_s} \sum_{j=1}^m \kappa(x_i,y_j)\sqrt{m} V_{js} \approx \sqrt{N} U_{is} \Rightarrow \frac{1}{\lambda_s} \sum_{j=1}^m   \kappa(x_i,y_j) V_{js} \approx  U_{is},
\end{aligned}
\end{equation}
which indeed correspond to matrix SVD in \eqref{eq:integral:svd:motivate}. 

Note that while this alternative above can also derive the asymmetric \nystrom method, it is different from the techniques presented in Section \ref{sec:nystrom}. In contrast,  the derivation in \eqref{sec:nystrom} starts from the integral equations of the pair of adjoint eigenfunctions with asymmetric kernels. One of our goals is to align and compare w.r.t. the symmetric \nystrom in \cite{NIPS2000_19de10ad}, which is widely adopted in machine learning, which views the \nystrom approximation \cite{bakerprenter1981numerical} originally from the integral equations with symmetric kernels, as presented in Section \ref{sec:supp:related:work:nystrom} in the Appendix, where thorough comparisons on the connections and differences are discussed.

\end{enumerate}

\section{Additional numerical results} \label{app:exp}

\subsection{Additional ablations on \ksvd}
To further study the effect of simultaneous nonlinearity and asymmetry in \ksvd, we design the following experiment.
We first make some non-linear encoding in a preprocessing step to the samples $x_i$ (i.e., rows of the given data matrix $A$) and then compute SVD, and compare the downstream classification/regression results with SVD on the asymmetric kernel matrix.
Specifically, we consider polynomial features with degree 2 of the samples $x_i$ as $\varphi(x_i)$ and then apply SVD to $\varphi(A)=[\varphi(x_1), \ldots, \varphi(x_N)]^\top$ as $\varphi(A) = U_A \Sigma_A V_A^\top$ and use $U_A$ as the learned embeddings.
Correspondingly, \ksvd employs the polynomial kernel of degree 2 $k_{\text{poly}}(x,z)=(x^\top z+1)^2$ and applies SVD to the asymmetric kernel matrix $G_{ij}=k(x_i, z_j)$ and we use the singular vectors $B_\phi$ as the learned embeddings for fair comparisons.
The embeddings are then fed to a linear classifier/regressor for the downstream classification/regression tasks as in \Cref{sec:general:data} in the main paper.

\begin{table}[ht!]
    \centering
    \caption{Ablation study on SVD applied after nonlinear preprocessing v.s. \ksvd.
    Higher values ($\uparrow$) are better for AUROC and lower values ($\downarrow$) are better for RMSE.} 
    \label{tab:mytable}
    \begin{tabular}{lcccccc}
    \toprule
    \multirow{2}{*}{Method} & \multicolumn{3}{c}{AUROC ($\uparrow$)}  & \multicolumn{3}{c}{RMSE ($\downarrow$)} \\ \cmidrule(lr){2-4} \cmidrule(lr){5-7}
    & Diabetes & Ionosphere & Liver & Cholesterol & Yacht  & Physicochemical-protein \\
    \midrule
    Nonlinear+SVD & 0.6296 & 0.7292 & 0.7032 & \textbf{49.0867} & 15.0002 & {5.9517}\\
    \ksvd & \textbf{0.7607} & \textbf{0.8374} & \textbf{0.7100} & 49.1592 & \textbf{14.6489} & {\textbf{5.4583}}\\
    \bottomrule
    \end{tabular}
\end{table}

This experiment shows the additional benefit brought by the construction on row space and column space, as $\mathcal{X}, \mathcal{Z}$ in our derivations,
and with the asymmetric kernel trick, 
instead of simply applying SVD to a matrix which is attained by applying some nonlinear transformation to the rows of the data matrix $A$.
In fact, our experiments show that \ksvd is an effective tool to learn more informative embeddings when the given data physically present asymmetric similarities as in \Cref{sec:test:graph,sec:Biclustering}  in the main paper, and it also shows 
better performance
for general datasets as experimented in \Cref{sec:general:data} in the main paper. 

\subsection{Additional results on the asymmetric \nystrom method}\label{sec:appendix:results:nystrom}
\begin{figure}[t!]
    \centering
    \begin{subfigure}[b]{0.3\textwidth}
        \centering
        \includegraphics{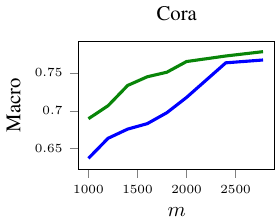}
        \label{fig:nystrom:macro:cora}
    \end{subfigure}
    \hfill
    \begin{subfigure}[b]{0.3\textwidth}
        \centering
        \includegraphics{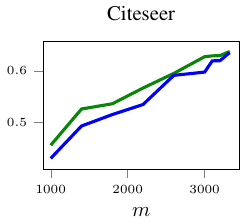}
        \label{fig:nystrom:macro:citeseer}
    \end{subfigure}
    \hfill
    \begin{subfigure}[b]{0.3\textwidth}
        \centering
        \includegraphics{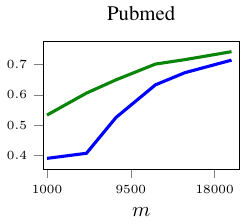}
        \label{fig:nystrom:macro:pubmed}
    \end{subfigure}

    \bigskip
    \begin{subfigure}[b]{0.3\textwidth}
        \centering
        \includegraphics{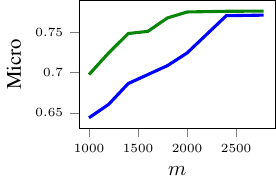}
        \label{fig:nystrom:micro:cora}
    \end{subfigure}
    \hfill
    \begin{subfigure}[b]{0.3\textwidth}
        \centering
        \includegraphics{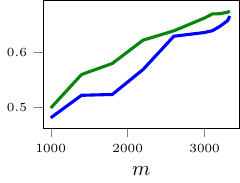}
        \label{fig:nystrom:micro:citeseer}
    \end{subfigure}
    \hfill
    \begin{subfigure}[b]{0.3\textwidth}
        \centering
        \includegraphics{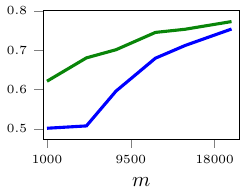}
        \label{fig:nystrom:micro:pubmed}
    \end{subfigure}
    \caption{F1 Scores at different numbers of subsampling $m$ with the asymmetric and symmetric \nystrom method. Green line: \ksvd, blue line: KPCA.}
    \label{fig:nystrom}
\end{figure}

In \Cref{fig:nystrom} in the main paper, the node classification F1 score is reported for multiple number of subsamplings $m$, where \ksvd (green line) employs the asymmetric \nystrom method and KPCA (blue line) uses the symmetric \nystrom, both employing the RBF kernel. 
Note that, as explained in the main paper, the resulting kernel matrix  $G$ in  \ksvd  maintains the asymmetry even with the (symmetric) RBF function, as the kernel is applied to two different inputs, i.e., $\mathcal X$ and $\mathcal Z$.
Note that the data matrix is square, so we can set $m=n$ for the subsamplings of the asymmetric \nystrom. In addition, we  provide the corresponding Micro F1 scores on Cora and also add the evaluations on Citeseer and Pubmed.
The asymmetric \nystrom-based kernel method \ksvd shows superior performances at all considered $m$ compared to KPCA without significant decrease in accuracy of the solution due to the subsampling. 

{In Table \ref{tab:speedup:low:high:appendix}, we provide extensional results on Table \ref{tab:speedup:low:high} for the tolerance levels $\varepsilon=10^{-1}$ and $10^{-2}$, showing the training time and the speedup w.r.t. RSVD, i.e. $t^\text{(RSVD)}/t^\text{(Ours)}$, where $t^\text{(RSVD)}, t^\text{(Ours)}$ is the training time of RSVD and our asymmetric \nystrom solver, respectively.
Our solver maintain  the fastest than the compared solvers and our improvement is more significant with larger problem sizes.}

\pgfplotsset{
    NystromF1/.style={
        width=0.99\linewidth,
        xlabel=$m$,
        height=3.75cm,
        legend pos=south east,
        legend cell align={left},
        xtick pos=left, ytick pos=left,
        xtick align=outside, ytick align=outside,
        every axis plot/.append style={no marks, line width=1.5pt},
        every axis legend/.code={\let\addlegendentry\relax},
        every tick label/.append style={font=\tiny},
    },
    NystromF1Cora/.style={
        NystromF1,
        xmin=900, xmax=2900,
    },
    NystromF1Citeseer/.style={
        NystromF1,
        xmin=900, xmax=3450,
    },
    NystromF1Pubmed/.style={
        NystromF1,
        xmin=600, xmax=20500,
        xtick={1000,9500,18000},
        scaled x ticks=false
    },
}

We further experiment on large-scale datasets with millions of samples and features  in Table \ref{tab:large:scale} below,  showing the classification performance (AUROC) of  KPCA/\ksvd with RBF with subsampling $m=1000$, where $N$  is the number of samples and $M$ is the number of variables. We employ alternative $a_2$ for the compatibility matrix $C$. In Table \ref{tab:large:scale}, \ksvd achieves the best performance also in real-world large datasets, further verifying the effectiveness and scalability.

  \begin{table*}[ht!]
        \caption{	
        Runtime for multiple \ksvd problems at {different} tolerances; the lowest solution time is in bold. 
        }
        \label{tab:speedup:low:high:appendix}
        \centering
        \resizebox{0.85\textwidth}{!}{
        \begin{tabular}{lccccccccccc}
            \toprule
            \multirow{2}{*}{Task} & \multirow{2}{*}{$N$} & \multicolumn{4}{c}{Time (s) for $\varepsilon=10^{-1}$}  & \multicolumn{1}{c}{} & \multicolumn{4}{c}{Time (s) for $\varepsilon=10^{-2}$} & \multicolumn{1}{c}{}  \\\cmidrule(lr){3-7} \cmidrule(lr){8-12}
            & & TSVD & RSVD & Sym.~Nys. & Ours & Speedup& TSVD & RSVD & Sym.~Nys. & Ours & Speedup\\	
            \midrule
            Cora       & 2708  & 0.841  &  0.274  & 0.673   & \textbf{0.160}  & 1.71$\times$  & 0.841  &  0.313  & 0.681   & \textbf{0.225}  & 1.39$\times$\\ 	
            Citeseer   & 3312  & 0.568  &  0.290  & 0.214   & \textbf{0.136}  & 2.14$\times$& 0.568  &  0.396  & 0.425   & \textbf{0.239}  & 1.66$\times$  \\ 	
            PubMed     & 19717 & 9.223  &  4.577  & 44.914  & \textbf{0.141}  & 32.51$\times$ & 9.223  &  5.209  & 53.297  & \textbf{0.590}  & 8.83$\times$ \\ 
            \bottomrule
        \end{tabular}
        }
    \end{table*}

\begin{table*}[ht!]
\caption{{Classification results with AUROC metric ($\uparrow$) on large-scale real-world datasets from \cite{chang2011libsvm}.}}\label{tab:large:scale}
    \centering
    \begin{tabular}{cccccc}
    \toprule
      Dataset   & $N$ & $M$ & KPCA &\ksvd  \\
      \midrule
       AmazonCat-13K&	1,186,239 &	203,882 & 	0.51 &	\textbf{0.55} \\
       Avazu &	14,596,137&	1,000,000	&0.52	&	\textbf{0.67} \\
Criteo	&45,840,617&	1,000,000&	0.54	&	\textbf{0.63} \\
       \bottomrule
    \end{tabular}
    \label{tab:my_label}
\end{table*}

\section{Experimental details} \label{app:setups}
Details of the experimental setups are provided below.
Experiments in \Cref{sec:test:graph,sec:Biclustering} are implemented in MATLAB 2023b, and {Python 3.7} is used in \Cref{sec:general:data}. 
Experiments are run on a PC with an Intel i7-8700K and 64GB RAM, and experiments in \Cref{sec:general:data} use a single NVIDIA GeForce RTX 2070 SUPER GPU.

\subsection{Feature learning experiments}
In the experiments, we conduct 10-fold cross validation for determining kernel hyperparameters with grid searches in the same range for fair comparisons. The employed nonlinear kernels in the experiments are $\hat{\kappa}_{\rm{RBF}}(  {x},  {z}) = \exp(-\frac{\|  {x}-  {z}\|_2^2}{\gamma^2})$ and $\kappa_{\rm{SNE}}(  {x},  {z}) = \frac{\exp(-\|  {x}-  {z}\|_2^2/\gamma^2)}{\sum_{  {z}'\in \mathcal{Z}}\exp(-\|  {x}-  {z}'\|_2^2/\gamma^2)}$ with hyperparameter $\gamma$.
In the node classification experiments, we denote $A =: X=[   x_1, \dots,    x_N]^\top$ as the asymmetric adjacency matrix with $X_{ij}$ as the directed similarity between node $i$ and node $j$. 
KPCA is conducted for feature extraction in the following way: we compute symmetric kernel matrix $\hat{G}$ s.t. $\hat{G}_{ij}=\hat{k}(   x_i,    x_j)$, with (symmetric) RBF kernel $\hat{k}$, and its top 1000 eigenvectors are taken as the extracted features taken as input to the LSSVM classifier, following \cite{he2022learning}. 
PCA is conducted similarly by taking the linear kernel $\hat{k}(   x_i,    x_j)=   x_i^\top    x_j$. 
For all methods, we employ an LSSVM classifier with regularization parameter set to 1 and we utilize the one-vs-rest scheme. 
We use the original implementations of the authors for all baselines and the best parameters reported in their papers.
\qh{Graph reconstruction is a typical task in node representation learning and  is helpful to evaluate how well the learned representations preserve neighborhood information in embedding space. Graph reconstruction reconstructs all existing edges by reconstructing the full adjacency matrix from embedding space.} In this task, with the  feature embeddings extracted by all tested methods, we recover the matrix that reflects the edges between nodes and then the connections between  each node. For a given node $v$ with the out-degree $k_v$, the  closest $k_v$ nodes to $v$ in feature space are searched to reconstruct the adjacency matrix. The $\ell_1, \ell_2$ norms between $X$ and its reconstruction are evaluated.

In biclustering tasks, the closely related baseline methods, i.e., SVD and  KPCA, are compared with \ksvd, where the kernel setup is the same as above. Specifically, we apply SVD and  \ksvd on the data matrix  with attained left and right singular vectors and then  $k$-means is adopted for performing the biclustering task  with extracted features, where we use the scikit-learn in Python to implement $k$-means. We note that as KPCA only works with symmetric kernels, so KPCA is applied twice.   We also compare 
with the  biclustering method EBC \cite{ebc} based on ensemble  and  
the recently proposed BCOT method \cite{fettal2022efficient} based on optimal transport. We follow the data setups and evaluations in \cite{fettal2022efficient} with official sources in \url{https://github.com/chakib401/BCOT}: the rows relate to the clustering of documents, where the ground truth can be compared through the popular clustering metric NMI; the columns relate to the clustering of terms, where the Coherence index is used
\cite{pmi}.
For the compatibility matrix $C$, we use alternative $a_1$. The  results of BCOT are taken from its orignal paper \cite{fettal2022efficient}, and  EBC are ran by its  official codes provided in \url{https://github.com/blpercha/ebc} with threshold  $10^{-4}$.
On the tested datasets, we provide their descriptions in Table \ref{tab:graph:data} and Table \ref{tab:biclustering:data}.
The  results of BCOT are from its paper \citep{fettal2022efficient}, {and  EBC are ran by official codes with threshold  $10^{-4}$.}

\begin{table}[t]
    \small
    \centering
    \caption{Descriptions of the tested directed graph datasets.} 
    \label{tab:graph:data}
    \begin{tabular}{lccccc}
    \toprule
    {Datasets} & {Cora} & {Citeseer} & {Pubmed} & TwitchPT & BlogCatalog \\
    \midrule
    \# Nodes & 2708 & 3327 & 19717 & 1,912 & 10,312\\
    \# Edges & 5429 & 4732& 44338 & 64,510 & 333,983\\ 
    \# Classes & 7 & 6 & 3 & 2 &  39 \\
    \bottomrule
    \end{tabular}
\end{table}

\begin{table}[t]
    \small
    \centering
    \caption{Descriptions of the tested datasets for biclustering \cite{fettal2022efficient}.} 
    \label{tab:biclustering:data}
    \begin{tabular}{lccccc}
    \toprule
    {Datasets} & ACM & DBLP & Pubmed & Wiki \\
    \midrule
    \# Documents & 3025 & 4057 & 19717& 2405 \\
    \# Terms & 1870 & 334 & 500 & 4973\\ 
    \# Document clusters & 3 & 4 & 3 & 17  \\
    \# Term clusters & 18 &2&3 &23\\
    \bottomrule
    \end{tabular}
\end{table}

\subsection{\nystrom experiments}
In this part, we evaluate the efficiency of the proposed asymmetric \nystrom method with comparisons to other standard solvers. 
The accuracy of a solution $\tilde{U}=[\tilde{   u}_1,\dots,\tilde{   u}_r], \tilde{V}=[\tilde{   v}_1,\dots,\tilde{   v}_r], $ is evaluated as the weighted average 
 $\eta = \frac{1}{r} \sum_{i=1}^r w_i ( 1 - |   u_i^\top \frac{\tilde{   u}_i}{\norm{\tilde{   u}_i}}| ) +
 \frac{1}{r} \sum_{i=1}^s w_i ( 1 - |   v_i^\top \frac{\tilde{   v}_i}{\norm{\tilde{   v}_i}}| )$,
with $w_i=\lambda_i$,
 where $r$ is the rank of the low-rank approximation, $U=[   u_1,\dots,   u_r], V=[   v_1,\dots,   v_r]$ are the left and right singular vectors of $G$ from its rank-$r$ compact SVD with singular values $\lambda_1 \geq \dots \geq \lambda_r$.
We compare our method with three common SVD solvers: truncated SVD (SVD) from the ARPACK library, Symmetric \nystrom \cite{NIPS2000_19de10ad} applied to $GG^\top$ and $G^\top G$, and randomized SVD (RSVD) \cite{halko2011}.
We employ the Lanczos method at rank $r$ \cite{lehoucq1998} for the SVD subproblem of symmetric \nystrom, and
we employ RSVD at rank $r$ for the SVD subproblem of asymmetric \nystrom.
Truncated SVD is run to machine precision for comparison.
For a given tolerance $\varepsilon$, we stop training when $\eta < \varepsilon$, with $\eta$ being the accuracy of a solution.

In \Cref{tab:speedup:low:high} in the paper, we evaluate multiple tolerances, i.e., $\varepsilon=10^{-1}, 10^{-2}$.
In particular, for RSVD, we increase the number of oversamples until the target tolerance is reached.
For the \nystrom methods, we increase the number of subsamples $m$ until the target tolerance is reached.
We use random subsampling for all \nystrom methods.
The tolerance used in \Cref{fig:spectrum,fig:nystrom:cora} is $\varepsilon=10^{-2}$.
In \Cref{tab:speedup:low:high}, the SNE kernel bandwidth is set as $\gamma=k\sqrt{M\gamma_x}$, with $\gamma_x$ the variance of the training data and data-dependent $k$ ($k=1$ for Cora and Citeseer, $k=0.5$ for Pubmed); e.g., for Cora  $\gamma_x=0.0002$ and $\gamma=k\sqrt{M\gamma_x}\approx 0.74$.
This gives an indication on the scaling w.r.t $\gamma$ in \Cref{fig:spectrum}.
In \Cref{fig:spectrum}, we consider that a solver's performance may depend on the singular spectrum of the kernel matrix, so we vary $\gamma$ as shown in the horizontal axis in \Cref{fig:spectrum}, where an increased $\gamma$ leads to spectra with faster decay, and assess training time. Our approach shows overall speedup compared to RSVD, and our asymmetric \nystrom requires significantly fewer subsamples on the matrices with faster decay of the singular spectrum, showing greater speedup w.r.t. RSVD in this scenario. 
In the experiments of Fig. \ref{fig:nystrom} in this Appendix and of \Cref{fig:nystrom:cora} in the main body, we compare the node classification performance of KPCA using symmetric \nystrom against \ksvd using our proposed asymmetric \nystrom. We use the RBF kernel for both KPCA and \ksvd, with $\gamma$ tuned via 10-fold cross validation. Note that \ksvd achieves higher performance at all considered subsamplings $m$, even if both methods use the RBF kernel. 
Similarly, even when symmetric kernel functions are chosen, the resulting $G$ matrix in the \ksvd solution w.r.t. \eqref{prop:dual} in the paper still maintains the asymmetry, as the two inputs of the kernel is applied to  $\mathcal X$ and $\mathcal Z$, respectively.

\subsection{General data experiments} \label{app:setups:generaldata}
In the experiments on general datasets, we consider three common classification datasets, including Diabetes of size 768, Ionosphere of size 351, Liver of size 583, and three commonly used regssion datasets, including Cholesterol of size 303, Yacht of size 308, and 
Physicochemical-protein of size 45730.
Note that, though only the embeddings for samples are needed in prediction, i.e., the {right singular vectors} in \ksvd and the eigenvectors in KPCA, the embeddings by \ksvd are  learned on an asymmetric kernel with two feature maps, while in KPCA they are learned with a symmetric kernel relating to a single feature map. 
To implement a learnable $C$ matrix in \ksvd, i.e., the alternative $a_3$ in Remark 3.2 in the paper, we utilize the backpropagation learning scheme with stochastic gradient descent (SGD) based optimizers for minimizing the loss in the downstream tasks. Correspondingly, we
set $C$ matrix as learnable parameters that can be backpropagated 
and optimized by SGD-based optimizer in an end-to-end manner.
To make $C$ learnable, we set $GV$ as the learned features on the data samples to the downstream classifier/regressor, where $V$ is chosen as  the top-4 right singular vectors of $G$. 
In this manner, gradient can be backpropagated, where $V$ is alternatively updated through the SVD on $G$.
To be specific, we adopt an iterative training scheme for conducting SVD on the asymmetric kernel matrix $G$ and updating other parameters:
\textit{i)} for input $X$ and $Z$, which is given as $Z:=X^\top C$ in this case, we compute the asymmetric kernel matrix $G:=[\kappa(  {x},  {z})]$, $  {x}\in X$, $  {z}\in Z$, and then conduct SVD on $G$ to obtain $V$ s.t. $GV=U\Lambda$.
\textit{ii)} As $C$ can only be backpropagated through $G$, we detach the gradient of $V$ computed in previous step
and fix it,
we then forward $X$, $Z$ 
to 
update $G$ and send the projected features of samples from \ksvd, i.e., $GV$, 
to the 
classification or regression head with the computed loss (cross-entropy loss or the mean squared error loss), and
update all the parameters 
except $V$. 
In other experiments using KPCA or fixed $C$, i.e., $a_0$, $a_1$, $a_2$, we also train these methods with SGD-based optimizers, which makes our \ksvd comparable to the learnable $C$ case in $a_3$ for fair and consistent evaluations. Here, 
the difference lies in that we only need to update the classification/regression head, as the projected features of all samples ($GV$) is fixed with the given input data.

We adopt SGD as the optimizer for the linear classification or regression head, where the learning rate is set to $10^{-3}$ for all experiments except Cholesterol ($10^{-1}$) and Physicochemical-protein ($10^{-4}$). We choose the first 4 right singular vectors, i.e., $GV_{[:,:4]}$, to feed forward to the classification or regression head. When RBF kernel is used,  $\gamma^2$ is selected as 1e7 in most cases except for Physicochemical-protein dataset, which is with 1e6. When SNE kernel is used,  $\gamma^2$ is selected as 1e5 in most cases except for Ionosphere dataset with 1e6, Liver dataset with 1e4. Moreover, since Physicochemical-protein is a larger dataset, we utilize batch-training mode where we fix the batch size to be 500. All experiments are run for 2000 iterations.

\section{Algorithm for $C$}
Algorithm \ref{alg:compat} details the realization of the compatibility matrix discussed in \Cref{sec:cce} in the main paper. 
Below, we consider the case $M > N$, where we construct the projection matrix $C_x \in \mathbb{R}^{M \times N}$ such that $XC_x \in \mathbb{R}^{N \times N}$. 
If $N > M$, we rather construct $C_z \in \mathbb{R}^{N \times M}$ such that $ZC_z \in \mathbb{R}^{M \times M}$. 
The construction of $C_z$ mirrors the algorithm for $C_x$ with the appropriate changes.
In the case of square matrix with $N=M$, $C=I_N$, with $I_N$ the identity matrix of size $N \times N$.

\begin{algorithm}[tb]
   \caption{Compatibility Matrix Realization.}
   \label{alg:compat}
\begin{algorithmic}
   \STATE {\bfseries Input:} $\mathcal{X}=\{   x_i \in \mathbb{R}^M \}_{i=1}^N$
   \STATE Define $X = [   x_1, \dots,    x_N]^\top$
   \IF{projection on $   x_i$}
        \STATE $C_x = \arg \min \limits_{ C} \norm{ X - X  C C^\top}^2_{\rm F}$  
        \COMMENT{Alternative $a_1$}
        \ELSIF{randomized projection} 
        \STATE $C_x = \text{randn}(M, N)$ \COMMENT{Alternative $a_2$}
        \ELSIF{pseudoinverse}
        \STATE $C_x =\left((XX^\top)^\dag X \right)^\top$  \COMMENT{Alternative $a_0$}
        \ELSIF{learnable}
        \STATE $C_x$ is learned  by optimizing the downstream task objective.
        \COMMENT{Alternative $a_3$}
        \ENDIF
        \STATE {\bfseries Return:} $C_x$

\end{algorithmic}
\end{algorithm}

\section{Proof of Proposition 2.2}
\begin{proof}
Let $B_\phi \in \R^{n \times r}$.
\begin{align*}
    [\Gamma_\psi \Gamma_\phi^* B_\phi]_{jl} &= \frac{1}{\sqrt{m}}\langle \psi(z_j) , \frac{1}{\sqrt{n}}\sum_{i=1}^n b^\phi_{il} \phi(x_i) \rangle \\
    &= \sum_{i=1}^n \frac{1}{\sqrt{nm}} \langle \phi(x_i), \psi(z_j) \rangle b^\phi_{il} \\
    &= [G^\top B_\phi]_{jl}
\end{align*}
The proof for $\Gamma_\phi \Gamma_\psi^*$ is similar.
\end{proof}
\section{Proof of Proposition 3.3}
\begin{proof}
Apply on the left respectively $\Gamma_{\phi}$ and $\Gamma_{\psi}$ to both equations from \Cref{eq:cce_gamma} combined with Proposition~3.2.
\end{proof}

\section{Proof of Proposition 3.4}
\begin{proof}
    Perform the substitution of the proposed $W_\phi, W_\psi$ in the CCE problem with the knowledge that $B^{\text{svd}}_\phi, B^{\text{svd}}_\psi$ come from the SVD of $G$.
\end{proof}